%% file: emnlp-ijcnlp-2019.tex
\documentclass[11pt,a4paper]{article}
\usepackage[hyperref]{emnlp-ijcnlp-2019}
\usepackage{times}
\usepackage{latexsym}

\usepackage{url}
\usepackage{enumitem}
\usepackage{microtype}
\usepackage{graphicx}
\usepackage{subfigure}
\usepackage{booktabs}
\usepackage{comment}
\usepackage{xcolor}
\usepackage{todonotes}
\usepackage{amsmath}
\usepackage{amsfonts}
\usepackage{amssymb}

\aclfinalcopy 

\title{Judge the Judges: A Large-Scale Evaluation Study of Neural \\Language Models for Online Review Generation}

\author{Cristina G\^arbacea$^{1}$, Samuel Carton$^{2}$, Shiyan Yan$^{2}$, Qiaozhu Mei$^{1,2}$ \\
$^{1}$Department of EECS, University of Michigan, Ann Arbor, MI, USA \\
$^{2}$School of Information, University of Michigan, Ann Arbor, MI, USA \\
 {\sf \{garbacea, scarton, shiyansi, qmei\}@umich.edu} \\
}

\date{}

\begin{document}
\maketitle
\input{abstract}
\input{introduction.tex}
\input{related_work}

\input{experiment_design}

\input{results}

\input{discussion}

\section*{Acknowledgement}

We thank Wei Ai for his help on the power analysis, and Yue Wang and Teng Ye for helpful discussions. This work is in part supported by the National Science Foundation under grant numbers 1633370 and 1620319 and by the National Library of Medicine under grant number 2R01LM010681-05.

\bibliography{emnlp-ijcnlp-2019}
\bibliographystyle{acl_natbib}

\input{appendix}

\end{document}

%% file: abstract.tex
\begin{abstract}

We conduct a large-scale, systematic study to evaluate the existing evaluation methods for natural language generation in the context of generating online product reviews. We compare human-based evaluators with a variety of automated evaluation procedures, including discriminative evaluators that measure how well machine-generated text can be distinguished from human-written text, as well as word overlap metrics that assess how similar the generated text compares to human-written references.  We determine to what extent these different evaluators agree on the ranking of a dozen of state-of-the-art generators for online product reviews.  We find that human evaluators do not correlate well with discriminative evaluators, leaving a bigger question of whether adversarial accuracy is the correct objective for natural language generation.  
In general, 
distinguishing machine-generated text is challenging even for human evaluators, and human decisions correlate better with lexical overlaps.  We find lexical diversity an intriguing metric that is indicative of the assessments of different evaluators. A post-experiment survey of participants provides insights into how to evaluate and improve the quality of natural language generation systems \footnote{ The experimental setup, data, and annotations are publicly available at: \url{https://github.com/Crista23/JudgeTheJudges}}.  

\end{abstract}

%% file: introduction.tex
\section{Introduction}
\label{sec:introduction}

Recent developments in neural language models \cite{mikolov2012context}, \cite{reiter2009investigation}, \cite{mikolov2011rnnlm}, \cite{mikolov2011extensions} have inspired the use of neural network based architectures for the task of natural language generation (NLG). 
 Despite fast development of algorithms, there is an urgency to fill the huge gap in evaluating NLG systems.  On one hand, a rigorous, efficient, and reproducible evaluation procedure is critical for the development of any machine learning technology and for correct interpretation of the state-of-the-art.  On the other hand, evaluating the quality of language generation is inherently difficult due to the special properties of text, such as \textit{subjectivity} and \textit{non-compositionality}. 
 Indeed, \textit{``there is no agreed objective criterion for comparing the goodness of texts''} \cite{dale1998towards}, and there lacks a clear model of text quality \cite{hardcastle2008can}. 


Conventionally, most NLG systems have been evaluated in a rather informal manner. \cite{reiter2009investigation} divide existing evaluation methods commonly employed in text generation into three categories: \textit{i)} evaluations based on task performance, 
\textit{ii)} human judgments and ratings, where human subjects are recruited to rate different dimensions of the generated texts, 
and \textit{iii)} evaluations based on comparison to a reference corpus using automated metrics. \textit{Task based evaluation} 
considers that the value of a piece of functional text lies in how well it serves the user to fulfill a specific application. 
It can be expensive, time-consuming, and often dependent on the good will of participants in the study. Besides that, it is hard to toss out the general quality of text generation from the special context (and confounds) of the task, or to generalize the evaluation conclusions across tasks. \textit{Human annotation} is able to assess the quality of text more directly than task based evaluation. However, rigorously evaluating NLG systems with real users can be expensive and time consuming, and it does not scale well \cite{reiter2001using}. Human assessments also need to be consistent and repeatable for a meaningful evaluation \cite{lopez2012putting}. Alternative strategies which are more effective in terms of cost and time are used more frequently.  

\textit{Automated evaluation} compares texts generated by the candidate algorithms to human-written texts.  Word overlap metrics and more recent automated adversarial evaluators are widely employed in NLG as they are cheap, quick, repeatable, and do not require human subjects when a reference corpus is already available.  In addition, they allow developers to make rapid changes to their systems and automatically tune parameters without human intervention.  Despite the benefits, however, the use of automated metrics in the field of NLG is controversial \cite{reiter2009investigation}, and their results are often criticized as not meaningful for a number of reasons.  First, these automatic evaluations rely on a high-quality corpus of references, which is not often available. 
Second, comparisons with a reference corpus do not assess the usefulness and the impact of the generated text on the readers as in human-based evaluations. 
Third, creating human written reference texts specifically for the purpose of evaluation could still be expensive, especially if these reference texts need to be created by skilled domain experts.  Finally and most importantly, using automated evaluation metrics is sensible only if they correlate with results of human-based evaluations and if they are accurate predictors of text quality, which is never formally verified at scale. 

We present a large-scale, systematic experiment that evaluates the \textit{evaluators} for NLG. We compare three types of evaluators including human evaluators, automated adversarial evaluators trained to distinguish human-written from machine-generated product reviews, and word overlap metrics (such as BLEU and ROUGE) in a particular scenario, generating online product reviews. The preferences of different evaluators on a dozen representative deep-learning based NLG algorithms are compared with human assessments of the quality of the generated reviews. Our findings reveal significant differences among the evaluators and shed light on the potential factors that contribute to these differences.  
The analysis of a post experiment survey also provides important implications on how to guide the development of new NLG algorithms. 




%% file: related_work.tex
\section{Related Work}
\label{sec:relatedwork}


\subsection{Deep Learning Based NLG}
\label{e2e}

Recently, a decent number of deep learning based models have been proposed for text generation.  
Recurrent Neural Networks (RNNs) and their variants, such as Long Short Term Memory (LSTM) \cite{hochreiter1997long} models, Google LM \cite{jozefowicz2016exploring}, and Scheduled Sampling (SS) \cite{bengio2015scheduled} are widely used for generating textual data.   

Generative Adversarial Networks \cite{goodfellow2014generative}, or GANs, train generative models through an adversarial process.  
Generating text with GANs is challenging due to the discrete nature of text data. SeqGAN \cite{yu2017seqgan} is one of the earliest GAN-based model for sequence generation, which treats the procedure as a sequential decision making process.  
RankGAN \cite{lin2017adversarial} proposes a framework that addresses the quality of a set of generated sequences collectively.   
Many GAN-based models \cite{yu2017seqgan}, \cite{lin2017adversarial}, \cite{rajeswar2017adversarial}, \cite{che2017maximum}, \cite{li2017adversarial}, \cite{zhang2017adversarial} are only capable of generating short texts.  LeakGAN \cite{guo2017long} is proposed for generating longer texts.  

Deep learning architectures other than LSTM or GAN have also been proposed for text generation. \cite{tang2016context} study NLG given particular contexts or situations and proposes two approaches 
based on the encoder-decoder framework. \cite{dong2017learning} address  the same task and employ an additional soft attention mechanism. 
Pre-training enables better generalization in deep neural networks \cite{erhan2010does}, especially when combined with supervised discriminative fine-tuning to learn universal robust representations \cite{radford2018improving}, \cite{devlin2018bert}, \cite{radford2019language}. 
\cite{guu2018generating} use a prototype-then-edit generative language model for sentences. 

\subsection{Automated Evaluation Metrics}
\label{automatic_eval}

The variety of NLG models are also evaluated with various approaches.  Arguably, the most natural way to evaluate the quality of a generator is to involve humans as judges, either through some type of Turing test \cite{machinery1950computing} to distinguish generated text from human input texts, or to directly compare the texts generated by different generators \cite{mellish1998evaluation}.  Such approaches are hard to scale and have to be redesigned whenever a new generator is included.  Practically, it is critical to find automated metrics to evaluate the quality of a generator independent of human judges or an exhaustive set of competing generators. 

\textbf{Perplexity} \cite{jelinek1977perplexity} is commonly used to evaluate the quality of a language model,  
 which has also been employed to evaluate generators \cite{yarats2017hierarchical}, \cite{ficler2017controlling}, \cite{gerz2018language} even though it is commonly criticized for not being a direct measure of the quality of generated text \cite{fedus2018maskgan}. Perplexity is a model dependent metric, and ``how likely a sentence is generated by a given model'' is not comparable across different models.  Therefore we do not include perplexity in this study. 

\textbf{Discriminative Evaluation} is an alternative way to evaluate a generator, which measures how likely its generated text can fool a classifier that aims to distinguish the generated text from human-written texts.  In a way, this is an automated approximation of the Turing test, where machine judges are used to replace human judges.  Discriminative machine judges can be trained either using a data set with explicit labels \cite{ott2011finding}, or using a mixture of text written by real humans and those generated by the model being evaluated.  The latter is usually referred to as \textit{adversarial evaluation}. \cite{bowman2015generating} proposes one of the earliest studies that uses adversarial error to assess the quality of generated sentences.  
Notably, maximizing the adversarial error is consistent to the objective of the generator in generative adversarial networks. 
\cite{kannan2017adversarial} propose an adversarial loss to discriminate a dialogue model's output from human output.  
The discriminator prefers longer output 
and rarer language instead of the common responses generated. 
 There however lacks evidence that a model that obtains a lower adversarial loss is better according to human evaluations. 
 
Automatic dialogue evaluation is formulated as a learning problem in \cite{lowe2017towards}, who train an RNN to predict the scores a human would assign to dialogue responses. RNN predictions correlate with human judgments at the utterance and system level, however each response is evaluated in a very specific context and the system requires substantial human judgments for training. \cite{li2017adversarial} employ a discriminator (analogous to the human evaluator in the Turing test) both in training and testing and define adversarial success. 
Other work finds the performance of a discriminative agent (e.g., attention-based bidirectional LSTM binary classifier) is comparable with human judges at distinguishing between real and fake dialogue excerpts \cite{bruni2017adversarial}. However, results show there is limited consensus among humans on what is considered as coherent dialogue passages. 

\textbf{Word Overlap Metrics}, such as BLEU \cite{papineni2002bleu}, ROUGE \cite{lin2004rouge}, and METEOR \cite{banerjee2005meteor}, are commonly used to measure the similarity between the generated text and human written references. \cite{liu2016not} find that word overlap metrics present weak or no correlation with human judgments in non-task oriented dialogue systems 
and thus should be used with caution or in combination with user studies.  In contrary, it is reported in \cite{sharma2017relevance} that text overlap metrics are indicative of human judgments in task-oriented dialogue settings, when used on datasets which contain multiple ground truth references. 
 \cite{dai2017towards} find text overlap metrics too restrictive as they focus on fidelity of wording instead of fidelity of semantics. 
 \cite{callison2006re} consider an increase in BLEU insufficient for an actual improvement in the quality of a system  
and posit in favor of human evaluation.

BLEU and its variants (e.g., Self-BLEU) are used to evaluate GAN models \cite{caccia2018language, zhu2018texygen}. 
\cite{shi2018towards} compare frameworks for text generation including MLE, SeqGAN, LeakGAN and Inverse Reinforcement Learning using a simulated Turing test. 
A benchmarking experiment with GAN models is conducted in \cite{lu2018neural}; results show LeakGAN presents the highest BLEU scores on the test data. Similarly, BLEU and METEOR present highest correlations with human judgements \cite{callison2008further}, \cite{graham2014re}. 
 However, evaluation metrics are not robust across conditions, and no single metric consistently outperforms other metrics across all correlation levels \cite{DBLP:journals/mt/PrzybockiPBS09}.

Conventional neural language models trained with maximum
likelihood can be on par or better than GANs \cite{caccia2018language}, \cite{semeniuta2018accurate}, \cite{tevet2018evaluating}. 
However, log-likelihood is often computationally intractable 
\cite{theis2015note}. Models with
good likelihood can produce bad samples, and vice-versa
\cite{goodfellow2016nips}. Generative models should be evaluated
with regards to the task they are intended
for over the full quality-diversity spectrum \cite{cifka2018eval}, \cite{hashimoto2019unifying}, \cite{montahaei2019jointly}. 

While many generators are proposed and evaluated with various metrics, no existing work has systematically evaluated the different evaluators at scale, especially in the context of online review generation.  Our work fills in this gap. 

%% file: experiment_design.tex
\section{Experiment Design}
\label{exp_design}

We design a large-scale experiment to systematically analyze the procedures and metrics used for evaluating NLG models. To test the different \textit{evaluators}, 
the experiment carefully chooses a particular application context and a variety of natural language generators in this context.  Ideally, a sound automated evaluator should be able to distinguish good generators from suboptimal ones.  Its preferences (on ordering the generators) should be consistent to humans in the exact application context. 


\subsection{Experiment Context and Procedure} We design the experiment in the context of generating online product reviews.  There are several reasons why review generation is a desirable task for the experiment: 1) online product reviews are widely available, and it is easy to collect a large number of examples for training/ testing the generators; 2) Internet users are used to reading online reviews, and it is easy to recruit capable human judges to assess the quality of reviews; and 3) comparing to tasks like image caption generation or dialogue systems, review generation has minimal dependency on the conversation context or on non-textual data, which reduces possible confounds. 

\begin{figure}[!htbp]
\centering
  \includegraphics[width=3in]{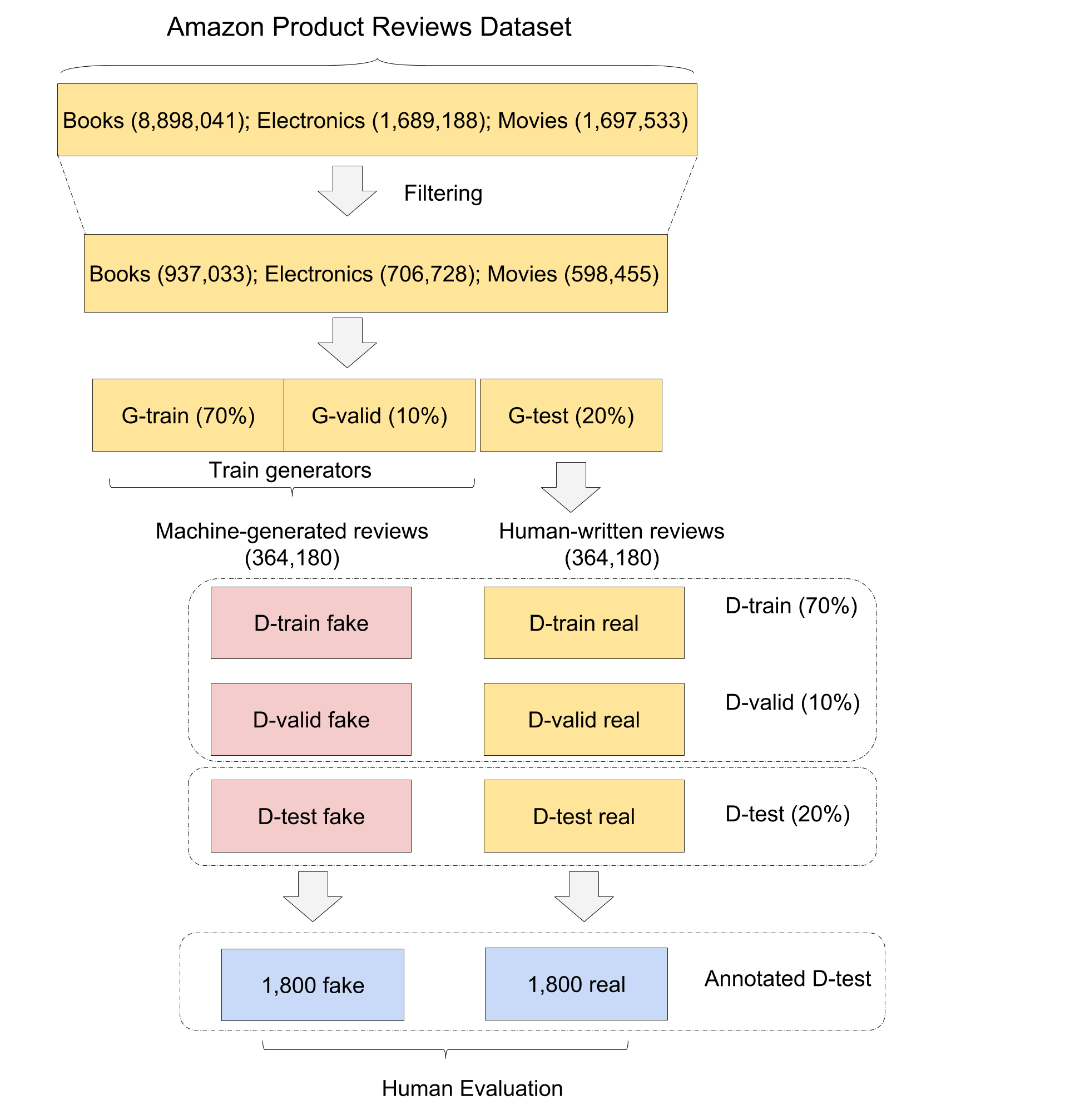}
  \caption{Overview of the Experiment Procedure.} 
  \label{dataset_split}
\end{figure}

The general experiment procedure is presented in Figure \ref{dataset_split}.  We start from the publicly available Amazon Product Reviews dataset \footnote{\url{http://jmcauley.ucsd.edu/data/amazon/}}
and select three most popular domains: \textit{books}, \textit{electronics}, and \textit{movies}.  
After filtering rare products, inactive users, and overly long reviews, the dataset is randomly split into three parts, to train, to validate, and to test the candidate review generators (denoted as \textit{G-train}, \textit{G-valid}, and \textit{G-test}).  Every generative model is trained and validated using the same datasets, and then charged to generate a number of product reviews (details are included in the next section).  These generated reviews, mixed with the real reviews in \textit{G-test}, are randomly split into three new subsets for training, validating, and testing candidate (discriminative) evaluators, denoted as \textit{D-train}, \textit{D-valid}, and \textit{D-test}. 
Finally, a random sample of reviews from \textit{D-test} are sent for human evaluation.



\subsection{Review Generators}
Although our goal is to evaluate the evaluators, it is critical to include a wide range of text generators with various degrees of quality. A good evaluator should be able to distinguish the high-quality generators from the low-quality ones.  We select a diverse set of generative models from recent literature.  The goal of this study is \textit{not} to name the best generative model, and it is unfeasible to include all existing models.  Our criteria are: (1) the models are published before 2018, when our experiment is conducted; (2) the models represent different learning strategies and quality levels; (3) the models have publicly available implementations, for reproducibility purposes.  In Table \ref{table:generators} we list the candidate generators.
It is not an exhaustive list of what are currently available. For implementation details of these models please see Appendix \ref{generators_implementation_details}.

\begin{table}[h]
\caption{Candidate models for review generation.}
\centering
\scalebox{0.65}{
\begin{tabular}{  l | c}
\hline
\hline
Generative Model & Adversarial  \\
 & Framework \\
\hline
\hline
Word LSTM temp 1.0 \cite{hochreiter1997long}  & No  \\
Word LSTM temp 0.7 \cite{hochreiter1997long} & No  \\
Word LSTM temp 0.5 \cite{hochreiter1997long} & No  \\
Scheduled Sampling \cite{bengio2015scheduled}  & No  \\
Google LM \cite{jozefowicz2016exploring} & No  \\
Attention Attribute to Sequence* \cite{dong2017learning} & No  \\
Contexts to Sequences* \cite{tang2016context} & No \\
Gated Contexts to Sequences* \cite{tang2016context}  & No \\
MLE SeqGAN \cite{yu2017seqgan} & Yes  \\
SeqGAN \cite{yu2017seqgan} & Yes \\
RankGAN \cite{lin2017adversarial} & Yes   \\
LeakGAN \cite{guo2017long} & Yes   \\
\hline
\hline
\end{tabular}}
\scriptsize * indicates that review generation using these models are conditional on context information such as product ids; other models are context independent. 
\label{table:generators}
\end{table}

Every generator (except Google LM) is trained and validated on \textit{G-train} and \textit{G-valid} datasets, and used to generate the same number of machine-generated (a.k.a., fake) reviews (see Table \ref{discriminator_student_dataset}).  We follow the best practice in literature to train these models, although it is possible that the performance of models might not be optimal due to various constraints.  This will not affect the validity of the experiment as our goal is to evaluate the \textbf{evaluators} instead of the individual generators.  Google LM was not trained on reviews, but it provides a sanity check for the experiment - a reasonable evaluator should not rank it higher than those trained for generating reviews. 

\begin{table}[!htbp]
\caption{Number of generated reviews by each model.} 
\centering
\scalebox{0.6}{
\begin{tabular}{ l | c | r | r  | r }
\hline
\textbf{Generative Model} & \textbf{Total} & \textbf{D-Train} & \textbf{D-Valid} & \textbf{D-Test}  \\
\hline
\hline
$\forall$ model in Table \ref{table:generators} except Google LM & 32,500 & 22,750 & 3,250 & 6,500  \\
Google LM & 6,680 & 4,676 & 668 & 1,336  \\
\hline
\hline
\end{tabular}}
\label{discriminator_student_dataset}
\end{table}

\subsection{Evaluators}
\label{eval_methods}

We include a comprehensive set of evaluators for the quality of the aforementioned generators: \textit{i)} human evaluators, \textit{ii)} discriminative evaluators, and \textit{iii)} text overlap evaluators.  The evaluators are the main subjects of the experiment. 

\subsubsection{Human evaluators} 
We conduct a careful power analysis  \cite{christensen2007methodology}, which 
suggests that at least 111 examples per generative model should be human annotated to infer that the machine-generated reviews are comparable in quality to human-written reviews, at a minimal statistically significance level of 0.05.  Per this calculation, we sample 150 examples for each of the 12 generators for human evaluation.  This totals 1,800 machine-generated reviews, to which we add 1,800 human-written reviews, or a total of 3,600 product reviews sent for human annotation. We markup out-of-vocabulary words in \textit{both} human-written and machine-generated reviews to control for confounds of using certain rare words. There is no significant difference in proportion of the markup token between the two classes (2.5\%-real vs. 2.2\%-fake). We recruit 900 human annotators through the Amazon Mechanical Turk (AMT) platform.    
Each annotator is presented 20 reviews, a mixture of 10 real (i.e., human written) and 10 fake (i.e., machine generated), and they are charged to label each review as real or fake based on their own judgment. Clear instructions are presented to the workers that markup tokens are present in both classes and cannot be used to decide whether a review is real or fake. 
Each page is annotated by 5 distinct human evaluators.  The 5 judgments on every review are used to assemble two distinct \textbf{human evaluators}: \textit{H1} - \textbf{individual votes}, treating all human annotations independently, and \textit{H2} - \textbf{majority votes} of the 5 human judgments. For every \textit{annotated} review,  the human evaluator ($H1$ or $H2$) makes a call which can be either right or wrong with regard to the ground truth.  
A generator is high quality if the human evaluator achieves low accuracy identifying the reviews as fake.  

\subsubsection{Discriminative evaluators} 
The inclusion of multiple generators provides the opportunity of creating \textbf{meta-adversarial evaluators}, trained using a \textit{pool} of generated reviews by \textit{many} generators, mixed with a larger number of ``real'' reviews (\textit{D-train} and \textit{D-valid} datasets).  Such a ``pooling'' strategy is similar to the standard practice used by the TREC conferences to evaluate different information retrieval systems \cite{harman2006trec}.  Comparing to individual adversarial evaluators, a meta-evaluator is supposed to be more robust and fair, and it can be applied to evaluate new generators without being retrained.  In our experiment, we find that the meta-adversarial evaluators rank the generators in similar orders to the best individual adversarial evaluators.  


We employ a total of 7 meta-adversarial evaluators: 3 deep, among which one using LSTM \cite{hochreiter1997long}, one using Convolutional Neural Network (CNN)  \cite{lecun1998gradient}, and one using a combination of LSTM and CNN architectures; 4 shallow, based on Naive Bayes (NB) \cite{rish2001empirical}, Random Forest (RF) \cite{liaw2002classification}, Support Vector Machines (SVM) \cite{cortes1995support}, and XGBoost \cite{chen2016xgboost}, with unigrams, bigrams, and trigrams as features and on balanced training sets.  
We find the best hyper-parameters using random search and prevent the models from overfitting by using early stopping. 
For every review in \textit{D-test} (either annotated or not), a meta-adversarial evaluator makes a judgment call.  A generator is considered high quality if the meta-adversarial evaluator makes more mistakes on reviews it generated. 

\subsubsection{Word-overlap evaluators} 

We include a set of 4 text-overlap metrics used for NLG evaluation: BLEU and METEOR (specific to machine translation), ROUGE (used in text summarization), and CIDEr \cite{vedantam2015cider} (used in image description evaluation). 
These metrics rely on matching $n$-grams in the target text (i.e., generated reviews) to the ``references'' (i.e., human-written reviews).  The higher the overlap (similarity), the higher the quality of generated text.  
For every generated review in \textit{D-test Fake}, we assemble the set of references by retrieving the top-$10$ most similar human-written reviews in \textit{D-test Real} using a simple vector space model. We compute 600-dimensional vector representation of reviews using Sent2Vec \cite{pagliardini2018unsupervised}, pretrained on English Wikipedia, and retrieve the top-k nearest neighbors for each review based on cosine similarity of the embedding vectors. The rationale of using nearest neighbors of each generated review as references is that to appear ``real'', a generated review just need to be similar to \textit{some} real reviews instead of \textit{all}.
A generator is considered high quality if its generated reviews obtain a high average score by a text overlap evaluator.  In total, we analyze and compare 13 candidate evaluators (2 human evaluators, 7 discriminative evaluators, and 4 text-overlap metrics), based on the \textit{D-test} dataset. 



%% file: results.tex
\section{Results}
\label{sec:results}

First, we are interested in the accuracy of individual evaluators - how well they can distinguish ``fake'' (machine-generated) reviews from ``real'' (human-written) reviews.  Second, we are interested in how an evaluator assesses the quality of the 12 generators instead of individual reviews.  
The absolute scores an evaluator gives to the generators are not as informative as how it ranks them: a good evaluator should be able to rank good generators above bad generators. 
Last but not least, we are interested in how the rankings by different evaluators correlate with each other.  Intuitively, an automated evaluator that ranks the generators in similar orders as the human evaluators is more reasonable and can potentially be used as the surrogate of humans.  

\subsection{Results of Individual Evaluators}
 
 
\subsubsection{Human evaluators}

Every review is annotated by 5 human judges as either ``fake'' or ``real.''  The inter-annotator agreement (Fleiss-Kappa score \cite{fleiss2013statistical}) is $k=0.2748$. 
This suggests that \textit{distinguishing machine-generated reviews from real in general is a hard task even for humans}; there is limited consensus on what counts as a realistic review. The low agreement also implies that any automated evaluator that mimics human judges is not necessarily the most ``accurate.''  

\begin{figure}[!h]
  \includegraphics[width=\linewidth, height = 2in]{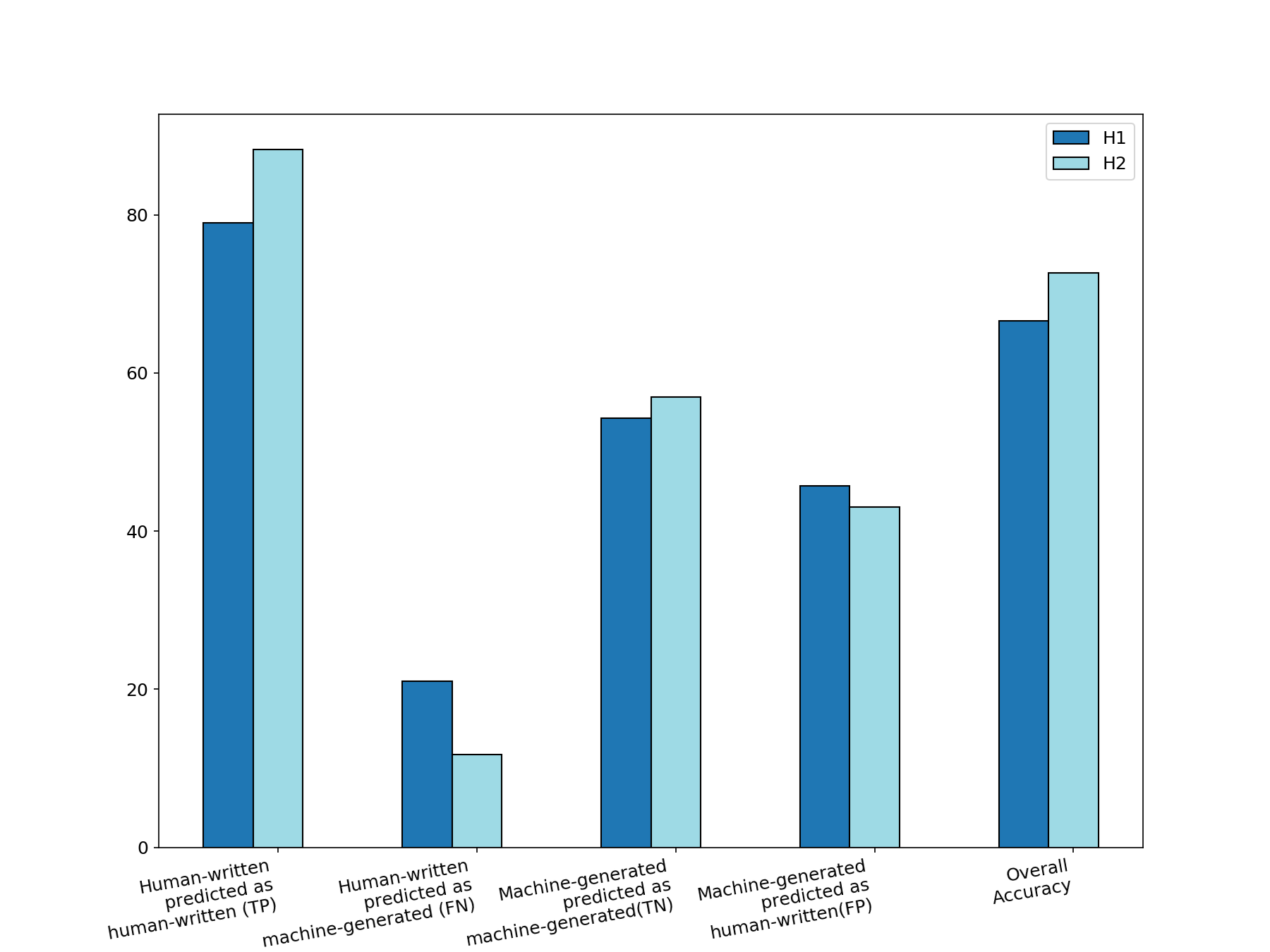}
  \small
  \caption{Accuracy of human evaluators on individual reviews: \textit{H1} - individual votes; \textit{H2} - majority votes. } 
  \label{fig::human_eval}
\end{figure}

In Figure \ref{fig::human_eval} we present the accuracy of two human evaluators on individual annotated reviews, based on either all 5 annotations or their majority votes for each review.  
Comparing to the ground-truth (of whether a review is machine-generated or collected from Amazon), individual human decisions are 66.61\% accurate, while their majority votes can reach 72.63\%.  Neither of them is close to perfect.  \textit{We observe that human evaluators generally do better at correctly labelling human-written reviews as real (true positive rate of 78.96\% for $H1$ and 88.31\% for $H2$), and they are confused by machine-generated reviews in close to half of the cases (true negative rate of 54.26\% for $H1$ and 56.95\% for $H2$)}.  This trend reassures previous observations \cite{tang2016context}. 

We then look at how the human evaluators rank the 12 generators, according to the accuracy of human evaluators on all (fake) reviews generated by each of the generators.   
The lower the accuracy, the more likely the human evaluator is confused by the generated reviews, and thus the better the generator.  
We observe a substantial variance in the accuracy of both human evaluators on different generators, which suggests that human evaluators are able to distinguish between generators. 
The generator ranked the highest by both human evaluators is \textit{Gated Contexts to Sequences}.  Google LM is ranked on the lower side, which makes sense as the model is not trained to generate reviews.  Interestingly, humans tend not to be fooled by reviews generated by the GAN-based models (MLE SeqGAN, SeqGAN, RankGAN and LeakGAN), even though their objective is to confuse fake from real.  GAN-generated reviews tend to be easily distinguishable from the real reviews by human judges.

\subsubsection{Discriminative evaluators}

We then analyze the 7 meta-adversarial evaluators. 
Different from human evaluators that are applied to the 3,600 annotated reviews, the discriminative evaluators are applied to \textit{all} reviews in \textit{D-test}. 

\textbf{Meta-adversarial Evaluators.} 
On individual reviews, the three deep learning based and the one SVM based evaluators achieve higher accuracy than the two human evaluators, indicating that adversarial evaluators can distinguish a single machine-generated review from human-written better than humans (Figure \ref{fig::H1_H2_LSTM_SVM}  and Table \ref{meta_discriminator_accuracy_rankings_all} in Appendix \ref{appendix_discriminative_evaluators_results}).  
Their true positive rates and true negative rates are more balanced than human evaluators. Meta-discriminators commonly rank GAN-based generators the highest.  
This makes sense as the objective of GAN is consistent to the (reversed) evaluator accuracy.  Interestingly, by simply setting the temperature parameter of Word LSTM to 1.0, it achieves comparable performance to the GANs.

\begin{figure}[!htbp]
\centering
  \includegraphics[width=\columnwidth]{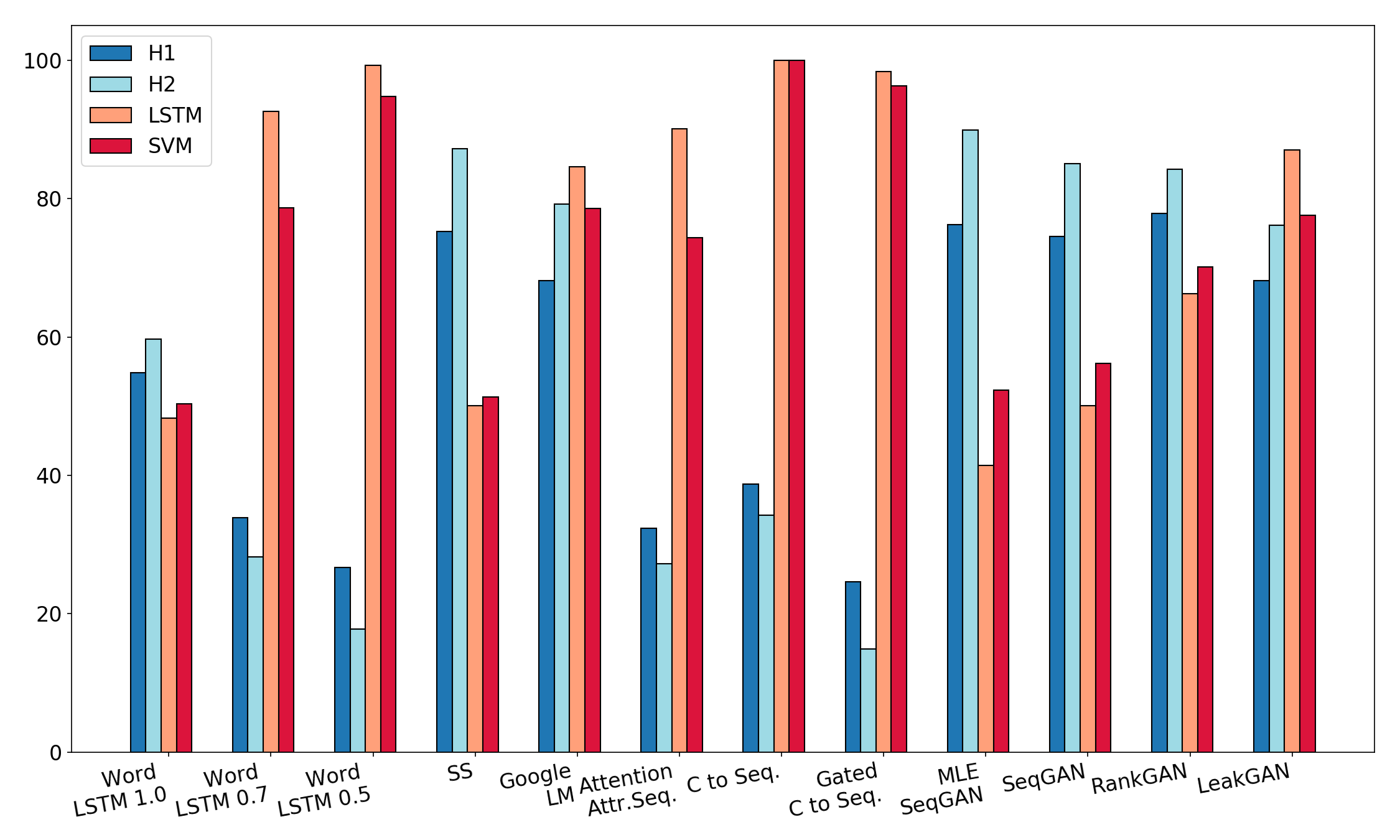}
  \caption{Accuracy of human (H1, H2) and meta-adversarial evaluators (LSTM, SVM) on reviews generated by individual generators. \textbf{The lower the accuracy, the better the generator.} }
  \label{fig::H1_H2_LSTM_SVM}
\end{figure}

\subsubsection{Word-Overlap Evaluators}

\begin{figure}[htbp]
\centering
  \includegraphics[width=\columnwidth]{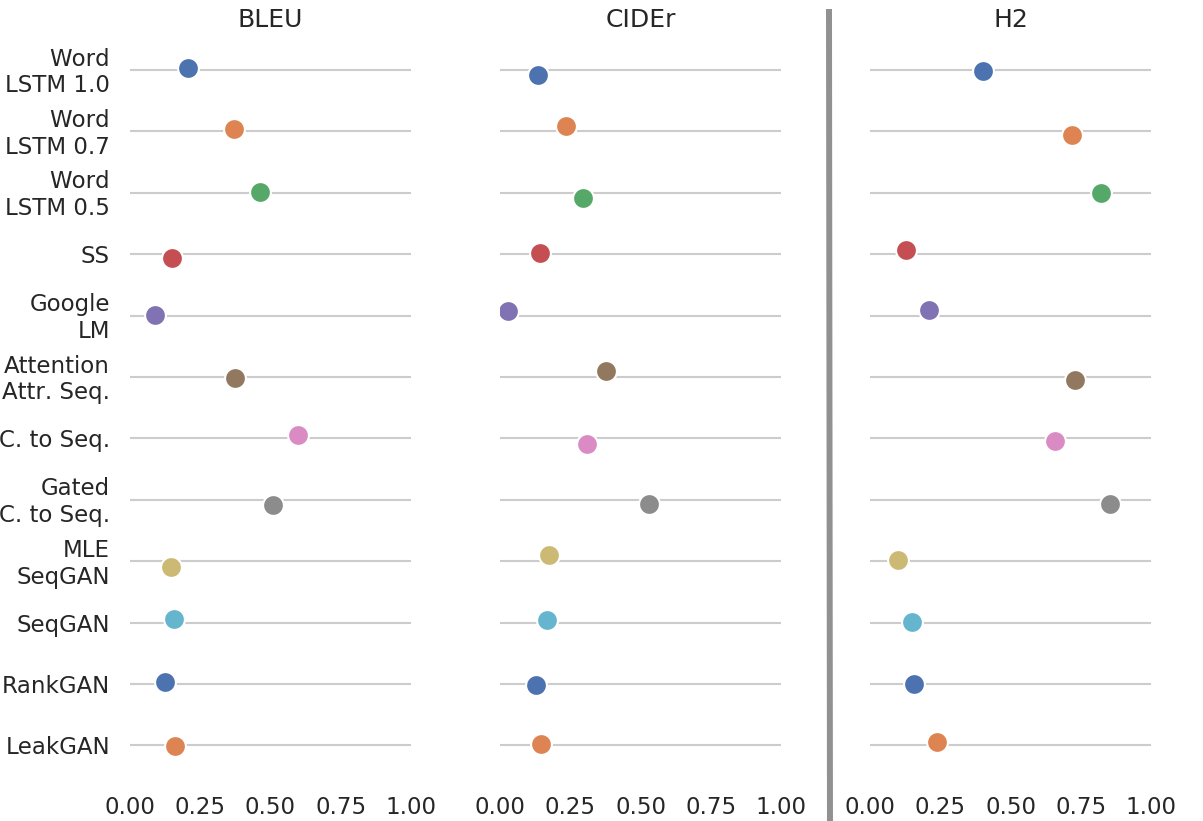}
  \caption{Text-Overlap Evaluators (BLEU and CIDEr) scores for individual generators. \textbf{The higher the better.}  The rankings are overall similar, as GAN-based generators are ranked low.}
  \label{word_overlap_eval}
\end{figure}

The generators are ranked based on the average scores of their generated reviews.  In Figure \ref{word_overlap_eval} we present the average scores of the 12 generators by each evaluator.  Different word-overlap evaluators also tend to rank the generators in similar orders. 
Interestingly, the top-ranked generator according to three evaluators is \textit{Contexts to Sequences}, while CIDEr scores highest the \textit{Gated Contexts to Sequences} model. GAN-based generators are generally ranked low; please also see Appendix \ref{appendix_text_overlap_evaluators_results}.


\subsection{Comparing Evaluators} 

To what degree do the evaluators agree on the ranking of generators? We are more interested in how the automated evaluators compare to the human evaluators, 
and whether there is any suitable automated surrogate for human judges at all. To do this, we compute the correlations between $H1$, $H2$ and each discriminative evaluator and correlations between $H1$, $H2$ and the text-overlap evaluators, based on either their decisions on individual reviews, their scores of the generators (by Pearson's coefficient \cite{fieller1957tests}), and their rankings of the generators (by Spearman's $\rho$ \cite{spearman1904proof} and Kendall's $\tau$ \cite{daniel1978applied}). 
Patterns of the three correlation metrics are similar; please see 
Figure~\ref{fig::barchart_correlation_evaluators} and Table \ref{table_correlation_results} in Appendix \ref{appendix_comparing_evaluators}. 

\begin{figure}[!h]
\centering
  \includegraphics[width=2.5in]{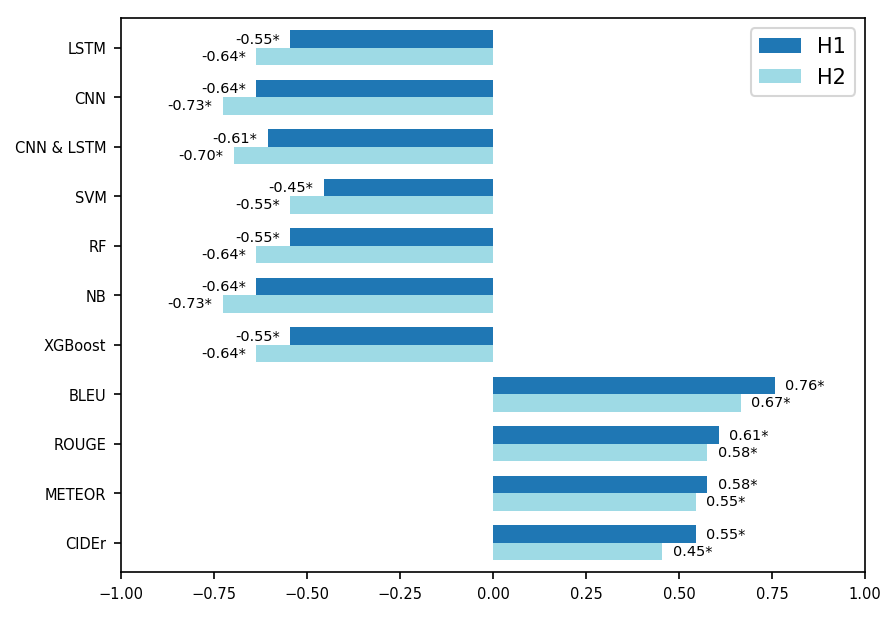}
  \caption{Kendall $\tau$-b between human and automated evaluators. Human's ranking is positively correlated to text-overlap evaluators and negatively correlated to adversarial evaluators ($^*$ is $p\le 0.05$).}
  \label{fig::barchart_correlation_evaluators}
\end{figure}

Surprisingly, none of the discriminative evaluators have a positive correlation with the human evaluators. 
That says, \textit{generators that fool machine judges easily are less likely to confuse human judges, and vice versa}. \textit{Word-overlap evaluators tend to have a positive correlation with the human evaluators in ranking the generators}.  Among them, BLEU appears to be closer to humans.  This pattern is consistent in all three types of correlations.  These two observations are intriguing, which indicates that when identifying fake reviews, humans might focus more on word usage rather than trying to construct a ``decision boundary'' mentally. 

In summary, we find that 1) human evaluators cannot distinguish machine-generated reviews from real reviews perfectly, with significant bias between the two classes; 2) meta-adversarial evaluators can better distinguish individual fake reviews, but their rankings at the generator level tend to be negatively correlated with human evaluators; and 3) text-overlap evaluators are highly correlated with human evaluators in ranking generators.  

%% file: discussion.tex
\section{Discussion}
\label{Discussion}

We carried a systematic experiment that evaluates the evaluators for NLG. Results have intriguing implications to both the evaluation and the construction of natural language generators. We conduct in-depth analysis to discover possible explanations.

\subsection{Granularity of Judgments}

We charged the Turkers to label individual reviews as either fake or real instead of evaluating each generator as a whole. Comparing to an adversarial discriminator, a human judge has not seen many ``training'' examples of \textit{fake} reviews or generators.  That explains why the meta-adversarial evaluators are better at identifying fake reviews. In this context, humans are likely to judge whether a review is real based on how ``similar'' it appears to the true reviews they are used to seeing online.

This finding provides interesting implications to the selection of evaluation methods for different tasks. In tasks that are set up to judge individual pieces of generated text (e.g., reviews, translations, summaries, captions, fake news) where there exists human-written ground-truth, it is better to use word-overlap metrics instead of adversarial evaluators. When judgments are made on the agent/ system level (e.g., whether a Twitter account is a bot), signals like how similar the agent outputs are or how much the agent memorizes the training examples may become more useful than word usage, and a discriminative evaluator may be more effective than word-overlap metrics. Our finding also implies that adversarial accuracy might not be the optimal objective for NLG if the goal is to generate documents that humans consider as real. Indeed, a fake review that fools humans does not necessarily need to fool a machine that has seen everything. In Appendix \ref{appendix_granularity_of_judgments} we provide more details.

\subsection{Imperfect Ground Truth}

One important thing to note is that all discriminative evaluators are trained using natural labels (i.e., treating all examples from the Amazon review dataset as positive and examples generated by the candidate models as negative) instead of human-annotated labels.  Some reviews posted on Amazon may have been generated by bots, and if that is the case, treating them as human-written examples may bias the discriminators.  To verify this, 
we apply the already trained meta-discriminators 
to the human-annotated subset (3,600 reviews) instead of the full \textit{D-test} set, and we use the majority vote of human judges (whether a review is fake or real) to surrogate the natural ``ground-truth'' labels (whether a review is generated or sampled from Amazon). 

\begin{figure}[!h]
\centering
  \includegraphics[width=2.5in]{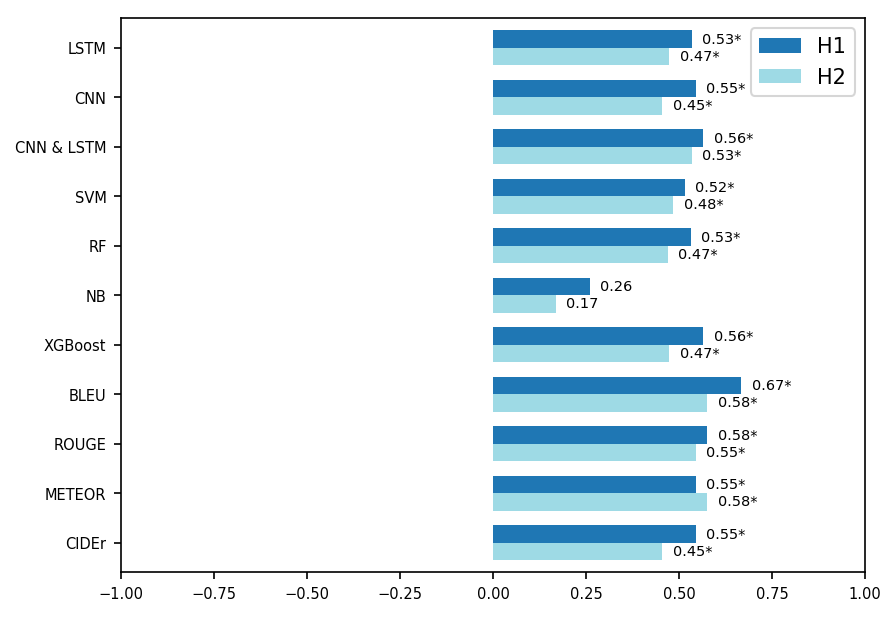}
  \caption{Kendall $\tau$-b correlation coefficients between human evaluators and automated evaluators, tested on the \textbf{annotated subset of \textit{D-test}} with \textit{majority votes} as ground-truth ($^*$ denotes $p \le 0.05$).}
  \label{fig::hbarchart_annotatedDtest_majority_vote_test_labels}
\end{figure}

When the meta-adversarial evaluators are tested using human majority-votes as ground-truth, the scores and rankings of these discriminative evaluators are more inline with the human evaluators, although still not as highly correlated as BLEU; please see Figure \ref{fig::hbarchart_annotatedDtest_majority_vote_test_labels}. Indeed, discriminative evaluators suffer the most from low-quality labels, as they were directly trained using the imperfect ground-truth.  Word-overlap evaluators are more robust, as they only take the most relevant parts of the test set as references (more likely to be high quality).  
Our results also suggest that when adversarial training is used, selection of training examples must be done with caution.  If the ``ground-truth'' is hijacked by low quality or ``fake'' examples, models trained by GAN may be significantly biased.  This finding is related to the recent literature of the robustness and security of machine learning models \cite{papernot2017practical}. 
Appendix \ref{appendix_imperfect_ground_truth} contains further details.

\subsection{Role of Diversity}
\begin{figure}[!htbp]
\centering
  \includegraphics[width=2.5in]{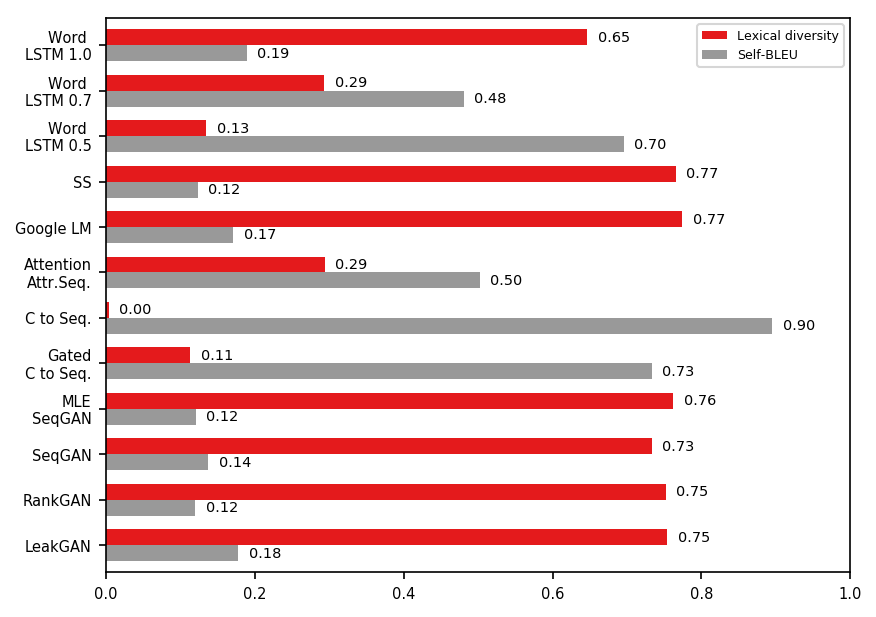}
  \caption{Self-BLEU scores (the lower the more diverse) and lexical diversity scores (the higher the more diverse) are highly correlated in ranking the generators.}
  \label{fig::self_bleu_lexical_diversity}
\end{figure}

We assess the role diversity plays in rankings the generators. Diversity of a generator is measured by either the lexical diversity \cite{bache2013text} or Self-BLEU \cite{zhu2018texygen} of the samples produced by the generator.  
Results obtained (see Figure \ref{fig::self_bleu_lexical_diversity}) indicate generators that produce the least diverse samples are easily distinguished by the meta-discriminators, while they confuse humans the most. This confirms that 
adversarial discriminators capture limitations of generative models in lack of diversity \cite{kannan2017adversarial}. 

Similarly, we measure to what extent the generators are memorizing the training set \textit{G-train} as the average BLEU scores of generated reviews using their nearest neighbors in \textit{G-train} as references.  We observe the generators do not memorize the training set, and GAN models generate reviews that have fewer overlap with \textit{G-train}; this finding is in line with recent theoretical studies on memorization in GANs \cite{theory19}. 


The effects of diversity indicate that when humans are distinguishing individual reviews as real or fake, whether or not a fake review is sufficiently different from the other generated reviews is not a major factor for their decision.  Instead, they tend to focus on whether the review looks similar to the reviews they have seen in reality.  A discriminative evaluator is more powerful in making decisions at a system level (e.g., a dialog system or a bot account), where diversity becomes a major factor. In Appendix \ref{appendix_role_of_diversity} we provide more details.


\subsection{User Study}

Finally, we are interested in the reasons why human annotators label certain reviews as fake (machine-written). After annotating a batch of reviews, workers are asked to explain their decisions by filling in an optional free-text comment.  This enables us to have a better understanding of what differentiates machine-generated from human-written reviews from human's perspective.  Analyzing their comments, we identify the main reasons why human evaluators annotate a review as machine-written.  These are mainly related to the presence of grammatical errors in the review text, wrong wording or inappropriate choice of expressions, redundant use of specific phrases or contradictory arguments in the review.  Interestingly, human evaluators' innate biases are also reflected in their decisions: they are likely to categorize a review as fake if it is too formal, lacks emotion and personal pronouns, or is too vague and generic.  
Please see Appendix \ref{appendix_user_study}.

\subsection{Summary}

In summary, our findings represent a preliminary foundation for proposing more solid and robust evaluation metrics and objectives of natural language generation. The low inter-rater agreement we observe suggests that judging \textit{individual} pieces of text as machine- or human-generated is a difficult task even for humans. In this context, discriminative evaluators are not as correlated with human judges as word-overlap evaluators. 
That implies that adversarial accuracy might not be the optimal objective for generating individual documents when realism is the main concern.  
In contrast, GAN based models may more easily pass a Turing test on a \textit{system} level, or in a conversational context.  When the judges have seen enough examples from the same generator, the next example had better be somewhat different. 

Our results also suggest that when adversarial evaluation is used, the training examples must be carefully selected to avoid false-positives.  We also find that when humans are distinguishing fake reviews from real ones, they tend to focus more on the usage of words, expressions, emotions, and other details. 
This may affect the design of objectives for the next generation of NLG models. 



%% file: appendix.tex
\appendix
\clearpage
\section{Appendix A}
\label{AppendixA}


\subsection{Implementation Details for Review Generators}
\label{generators_implementation_details}


\textit{Recurrent Neural Networks} (RNNs) directly model the generation process of text sequences, and provide an end-to-end solution to learning the generating function from large quantities of data. These networks maintain a hidden layer of neurons with recurrent connections to their own previous values, which in theory gives them the potential to model long span dependencies. For an input sequence $x = x_1, x_2, \ldots, x_t$, the hidden state $h_t$ which summarizes the information of the entire sequence up to timestep $t$  is recursively updated as $h_t = f(h_{t-1}, x_t)$, where $f(.,.)$ denotes a non-linear transformation function. The overall probability of the sequence is calculated as:

\begin{equation}
p(x) = \prod_{t=1}^{T}p(x_t|h_{t-1}),
\end{equation}
and the probability of generating the next word $x_{t+1}$ given its low dimensional continuous representation $O_{x_{t+1}}$ and input sequence $x_t$ is defined as:

\begin{equation}
p(x_{t+1} | x \le t) = p(x_{t+1}|h_t) \propto \exp(O_{x_{t+1}}^T h_t)
\end{equation}
However, in practice the gradient computation is difficult to propagate back in time due to exploding or vanishing gradients \cite{hochreiter2001gradient}, \cite{bengio1994learning}, making the learning of arbitrarily long phenomena challenging in RNNs. Long Short Term Memory networks (LSTMs) \cite{hochreiter1997long}  effectively address these limitations by relying on a memory state and gating functions to control the flow of the information throughout the network -- and in particular what information is written to the memory state, what information is read from the memory state, and what information is removed (or forgotten) from the memory state. The mathematical formulation of LSTM units can be expressed as follows:

\begin{equation}
\label{lstm_equations}
\begin{split}
i^{(t)} &= \sigma(W^{(i)}x^{(t)} + U^{(i)}h^{(t-1)} ) \qquad \text{(Input gate)} \\
f^{(t)} &= \sigma(W^{(f)}x^{(t)} + U^{(f)}h^{(t-1)}) \qquad \text{(Forget gate)} \\
o^{(t)} &= \sigma(W^{(o)} x^{(t)} + U^{(o)} h^{(t-1)}) \qquad \text{(Output gate)} \\
\widetilde{c}^{(t)} &= \text{tanh}(W^{(c)} x^{(t)} + U^{(c)}h^{(t-1)} \qquad \text{(New memory cell)} \\
c^{(t)} &= f^{(t)} \circ \widetilde{c}^{(t-1)} + i^{(t)} \circ \widetilde{c}^{(t)} \qquad \text{(Final memory cell)} \\
h^{(t)} &= o^{(t)} \circ \text{tanh}(c^{(t)}) \\
\end{split}
\end{equation}
In the above set of equations, the input word $x^{(t)}$ and the past hidden state $h^{(t-1)}$ are used to generate new memory $\widetilde{c}^{(t)}$ which includes features of the new word $x^{(t)}$ without prior determination of whether $x^{(t)}$ is important and worth keeping. The role of the input gate is to check whether it is sensible to store the new input word given the word $x^{(t)}$ itself and the past hidden state $h^{(t-1)}$; the input gate produces $i^{(t)}$ as output, which encapsulates the worthiness decision of preserving the input information. Similarly to the input gate, the forget gate also determines the usefulness of a word by inferring whether the past memory cell is used to compute the current memory cell by looking at the input word word $x^{(t)}$ itself and the past hidden state $h^{(t-1)}$; it produces $f^{(t)}$ as output, which encapsulates the worthiness decision of preserving the past memory cell. In the final memory generation stage, the advice of the input gate $i^{(t)}$ to gate the new memory $\widetilde{c}^{(t)}$ and the advice of the forget gate $f^{(t)}$ to forget the past memory $\widetilde{c}^{(t-1)}$ are both considered, and the two results are summed up to produce the final memory $c^{(t)}$. The output gate is used to separate the hidden state $h^{t}$ from the final memory of the network $c^{(t)}$. Given that every state of the LSTM is relying on hidden states and that the final memory $c^{(t)}$ contains a lot of information not necessarily required to be saved in the hidden state, the output gate discriminatively assesses which parts of the memory $c^{(t)}$ should be kept inside the hidden state $h^{t}$. In our experiments we employ an LSTM  generative model trained at word level. Sampling from a trained word language model can be done in two ways: beam search \cite{bahdanau2014neural} and random sampling \cite{graves2013generating}. Following \cite{tang2016context}, we use random sampling with different values for the temperature parameter. Sampling from the LSTM model with a high temperature results in the model generating diverse samples at the cost of introducing some mistakes, while small temperatures generate conservative samples without a lot of content diversity. In our experiments, we empirically set the temperatures to the following values: 1.0, 0.7 and 0.5.



RNNs, and LSTMs in particular, have become the standard for modeling machine learning problems that involve temporal and sequential data including text. The data is modeled via a fully-observed directed graphical model, where the distribution over a discrete time sequence $y_1, y_2, \dots, y_T$ is decomposed into an ordered product of conditional distributions over tokens:

\begin{equation}
P(y_1, y_2, \dots, y_T) = P(y_1)\prod_{t=1}^{T}P(y_t|y_1, \dots, y_{t-1})
\end{equation}
For models with recurrent connections from their outputs leading back into the model, \textit{teacher forcing} \cite{williams1989learning} is the most popular training strategy. This procedure emerges from the maximum likelihood criterion, in which at training time $t+1$ the model receives as input the ground truth output $y^t$:

\begin{equation}
\label{maxlikelihood}
\begin{split}
\log p(y^{(1)}, y^{(2)} | x^{(1)}, x^{(2)}) &= \log p (y^{(2)}| y^{(1)}, x^{(1)}, x^{(2)}) \\
&+ \log p (y^{(1)} | x^{(1)}, x^{(2)})
\end{split}
\end{equation}
The model in Equation \ref{maxlikelihood} above illustrates the conditional maximum likelihood criterion at timestep $t=2$. The model is trained to maximize the conditional probability of $y^{(2)}$ given the sequence $x$ generated so far and the previous $y^{(1)}$ value. Therefore, maximum likelihood specifies that at training time the previous token generated by the model is replaced with ground-truth examples $y_t$ that are fed back into the model for predicting outputs at later time steps. Feeding back ground truth samples at training time forces the RNN to stay close to the ground-truth sequence. However, at inference time, the ground truth sequence is no longer available conditioning, and each $y_t$ is generated by the model itself (i.e. sampled from its conditional distribution over the sequence given the previously generated samples). This discrepancy between training time and inference time causes errors in the model predictions that accumulate and amplify quickly over the generated sequence as the model is in a part of the state space it has never seen during training time. Small prediction errors compound in the RNN's conditioning context, and as the generated sample starts to diverge from sequences it has seen during training, the prediction performance of the RNN worsens \cite{lamb2016professor}. 
  



 
To alleviate this problem, Bengio et al  \cite{bengio2015scheduled} propose \textit{Scheduled Sampling (SS)}, a learning strategy for training RNNs which mixes inputs from the ground-truth sequence with inputs generated by the model itself at training time. SS relies on curriculum learning \cite{bengio2009curriculum} to change the training process from a fully guided scheme using the true previous token to a less guided scheme mostly using the generated token. The choice of replacing the ground truth with the model's prediction is determined by a coin flip with some probability, independently for each token. The probability of using the ground truth is set to a high value initially. As the model gradually keeps improving, samples from the model become more frequent and the model is partially fed with its own synthetic data as prefix  in a similar way to inference mode. Therefore, the training objective is slowly changed from an easy task where the previous token is known, to a realistic task where the previous token is provided by the model itself. The scheduled sampling training scheme is meant to make the model more robust and forces it to deal with its own mistakes at training time, in a similar way to inference time. However, as the model generates several consecutive tokens $y_t$-s, it is not clear whether the correct target distribution remains the same as in the ground truth sequence. The authors propose two solutions: \textit{i)} make the self-generated sequences short, and \textit{ii)} anneal the probability of using self-generated vs. ground-truth samples to 0, according to some schedule. 

Despite its impressive empirical performance, Huszar et al \cite{huszar2015not} show that SS is an inconsistent training strategy which pushes models towards memorising the distribution of symbols conditioned on their position in the sequence instead of on the prefix of preceding symbols. According to the authors, SS pays no attention to the content of the sequence prefix, and uses the hidden states to implement a simple counter which makes the model likely to recover from its own mistakes. Moreover, it is possible that the good performance of the model on image captioning datasets is either due to the algorithm not running until convergence, or to a lucky combination of factors including the model structure, early stopping, random restarts, and the annealing schedule. The authors recommend adversarial training strategies as a much better choice for generative models.





Tang et al \cite{tang2016context} study the the problem of NLG at particular contexts or situations. The authors focus on user review data due to its richness of context, sentiments and opinions expressed. They propose two approaches built on top of the encoder-decoder framework to generate user reviews as text sequences from user product contexts. In the first approach, \textit{Contexts to Sequences}, the authors encode the product context information $\overrightarrow{C}=\{\overrightarrow{c}_i\}_{i=1,\ldots, K}$, where $\overrightarrow{c}_i$ denotes a type of context and $K$ the number of context types, into a continuous semantic representation, which is fed into an LSTM decoder to generate text sequences. Despite promising results shown by the method, the authors consider that for long generated sequences the information from contexts is not propagated to distant words. In their second approach, \textit{Gated Contexts to Sequences}, the authors add skip-connections to directly build the dependency between contexts $h_C$ and each word when predicting the next word $x_{t+1}$ in a sequence. When a new word in a sequence is generated, it does not only depend on the current hidden state $h_t$, but it also depends on the context representation $h_C$. Similar to the first model, the decoder is a vanilla recurrent neural network with LSTM unit.

Focusing on the same problem as Tang et al \cite{tang2016context}, Dong et al \cite{dong2017learning} propose  \textit{Attention Enhanced Attribute to Sequence Model}. The model learns to encode product attributes into vectors by means of an encoder network,  and then generate reviews by conditioning on the encoded vectors inside a sequence decoder, and an attention mechanism \cite{bahdanau2014neural}, \cite{xu2015show} which learns soft alignments between the input attributes and the generated words. The product review generation problem is formally defined as follows. Given input attributes $a=(a_1, \ldots, a_{|a|})$, generate a product review $r=(y_1, \ldots, y_{|r|})$ which maximizes the conditional probability $p(r|a)$:

\begin{equation}
p(r|a) = \prod_{t=1}^{|r|}p(y_t| (y_1, \ldots, y_{t-1}), a)
\end{equation}
While the number of attributes $|a|$ is fixed for each product, the review text $r$ is a sequence of variable length. In our experiments we use the two models proposed by Tang et al \cite{tang2016context} and Dong et al \cite{dong2017learning} to generate use product reviews given the context information and the review text of each product in the Amazon dataset.
 

In addition to the already mentioned models, we also employ a pre-trained model released by Google, commonly referred to as Google LM \cite{jozefowicz2016exploring}. The model is an important contribution to the field of neural language modeling which emphasizes large scale recurrent neural network training. The model was trained on the One Billion Word Benchmark \cite{chelba2013one}, a publicly available dataset containing mainly news data and used as a reference standard for measuring the progress of statistical language modeling. The dataset includes 1 billion words in total with a vocabulary of 800,000 unique words. While for count based language models it is considered a medium-sized dataset, for neural network based language models the benchmark is regarded as a very large dataset. In terms of the model architecture, the GoogleLM model is a 2-layer LSTM neural network with 8,192 and respectively 1,024 hidden units in each layer, the largest Google was able to fit into GPU memory. The model uses Convolutional Neural Networks (CNNs) character embeddings as input, and makes predictions one character at a time, which presents the advantage that the model does not need to learn long-term dependencies in the data. We employ GoogleLM to generate sentences with a topic which identifies with the existing three categories (books, electronics and movies) present in the Amazon dataset we used.

Generative Adversarial Networks (GANs) \cite{goodfellow2014generative}   represent a training methodology for generative models via an adversarial process, and are aimed at generating synthetic data which resembles the real data. The GAN framework works through the interplay between two feedforward neural network models, a generative model $G$ and a discriminative model $D$, trained simultaneously by competing against each other. The generative model $G$ aims to capture the data distribution and generate high quality synthetic data, while the discriminative model $D$ estimates the probability a sample comes from the real training data and not from the synthetic data generated by $G$. Concretely, the generator $G$ takes as input a vector of random numbers $z$, and transforms it into the form of the data we are interested in imitating; the discriminator $D$ takes as input either the real data $x$ or generated data $G(z)$, and outputs probability $P(x)$ of the respective data being real. The GAN framework is equivalent to a minimax two-player game between the two models $G$ and $D$:

\begin{equation}
\label{GAN_equation}
\begin{split}
\min_G \max_D V(D, G) &= \mathbb{E}_{x \sim p_{\text{data}}(x)} [\log D(x)] \\
&+ \mathbb{E}_{z \sim p_z(z)}[\log(1-D(G(z)))]
\end{split}
\end{equation}

Adversarial learning algorithms iteratively sample batches from the data and noise distributions, and use noisy gradient information to simulatenously ascend in the parameters $\theta_d$ of $D$, while descending in the parameters $\theta_g$ of $G$. The discriminator $D$ is optimized to increase the likelihood of assigning a high probability to the real data $x$ and a low probability to the fake generated data $G(z)$. The gradient for the discriminator can be expressed as follows:

\begin{equation}
\label{discriminator_optimization}
\triangledown_{\theta_d} \frac{1}{m} \sum_{i=1}^{m} \big[\log D (x^{(i)}) + \log(1-D(G(z^{(i)})))\big]
\end{equation}

Alternatively, the generator $G$ is optimized to increase the probability the generated data $G(z)$ is rated highly:

\begin{equation}
\label{generator_optimization}
\triangledown_{\theta_g} \frac{1}{m} \sum_{i=1}^{m} \big[\log(1-D(G(z^{(i)})))\big]
\end{equation}

The goal of the generator $G$ is to maximize the probability of discriminator $D$ making a mistake by generating highly realistic data, while the discriminator $D$ is learnt to distinguish whether a given data instance is real or not. The gradient of the training loss from the discriminator $D$ is used as guidance for updating the parameters of the generator $G$. Gradient optimization is alternated between the two networks $D$ and $G$ as illustrated in Equations \ref{discriminator_optimization} and \ref{generator_optimization} on batches of real and generated data until GAN converges, at which point the data produced by GAN is the most realistic the network is capable of modeling.


However, GAN's applicability to discrete data is limited, despite the great success at generating realistic real valued synthetic samples in many computer vision tasks for eg., image generation \cite{brock2016neural}, \cite{zhu2016generative}, \cite{taigman2016unsupervised}, image style transfer \cite{luan2017deep}, \cite{zhu2017unpaired} and semantic segmentation \cite{luc2016semantic}, \cite{souly2017semi}. Training generative models of text using GANs is challenging due to the discrete nature of text data, which makes it difficult to backpropagate the gradient from the discriminator $D$ to the generator $G$. GANs are designed for generating real-valued, continuous data, and the gradient of the loss from discriminator $D$ w.r.t. the output of generator $G$ is used to guide $G$ to slightly change the generated value to make it more realistic (i.e. the gradient of the output of the discriminator network with respect to the synthetic data indicates how to slightly change the synthetic data to make it more plausible). Changes can be made to the synthetic data if it is based on real numbers, however for discrete tokens the slight change guidance is not a useful signal, as it is very likely that there is no corresponding token to the slight change given the limited vocabulary space\footnote{\url{https://www.reddit.com/r/MachineLearning/comments/40ldq6/generative_adversarial_networks_for_text/}}. In addition, a further reason why GANs cannot be applied to text data is because the discriminator $D$ can only asses a complete sequence. When having to provide feedback for partially generated sequences, it is non-trivial to balance the current score of the partially generated sequence with the future score after the entire sequence has been generated \cite{yu2017seqgan}. In the literature there are two approaches on how to deal with the problem of non-differentiable output and finding the optimal weights in a neural network: the REINFORCE algorithm, and Gumbel-Softmax reparameterization. We present  each method below.

\textit{REINFORCE}  \cite{williams1992simple} algorithms, also known as \textit{REward Increments, score-function estimators}, or \textit{likelihood-ratio methods} adjust the weights of a neural network based on the log derivative trick in a direction that lies along the gradient of expected reinforcement without explicitly computing gradient estimates. It is a policy gradient method which uses the likelihood ratio trick $\big(\frac{\triangledown_\theta p(X, \theta)}{P(X, \theta)} = \triangledown_{\theta} \log p(X, \theta); \frac{\partial}{\partial_x} \log f(x)=\frac{f'(x)}{f(x)} \big)$ to update the parameters of an agent and increase the probability that the agent's policy will select a rewarding action given a state. Given the trajectory $\tau_t = (u_1, \ldots, u_{t-1}, x_0, \ldots, x_t)$ made up of a sequence of states $x_k$ and control actions $u_k$, the goal of policy gradient is to find policy $\pi_{\vartheta}$ which takes as input trajectory $\tau_t$ and outputs a new control action that maximizes the total reward after $L$ time steps. $\pi_{\vartheta}$ is a parametric randomized policy which assumes a probability distribution over actions:




\begin{equation}
p(\tau; \vartheta) = \prod_{t=0}^{L-1}p(x_{t+1} | x_t, u_t) \pi_v(u_t|\tau_t)
\end{equation}
If we define the reward of a trajectory as:

\begin{equation}
R(\tau) = \sum_{t=0}^{N} R_{t}(x_t, u_t),
\end{equation}
the reinforcement learning optimization problem becomes:

\begin{equation}
\begin{split}
\max_{\vartheta}  J(\vartheta) = \max_{\vartheta}  \mathbb{E}_{p(\tau|\vartheta)} [R(\tau)]
\end{split}
\end{equation}
Then policy gradient can be derived as follows:

\begin{equation}
\label{policy_gradient}
\begin{split}
\triangledown_{\vartheta} J(\vartheta) &= \int R(\tau) \triangledown_{\vartheta} p(\tau; \vartheta)d\tau \\
&= \int R(\tau) \frac{\triangledown_{\vartheta} p(\tau; \vartheta)}{p(\tau; \vartheta)}p(\tau; \vartheta)d\tau  \\
&= \int (R(\tau)\triangledown_{\vartheta} \log p(\tau; \vartheta))p(\tau; \vartheta) d\tau   \\
&= \mathbb{E}_{p(\tau;\vartheta)} [R(\tau) \triangledown_{\vartheta} \log p(\tau; \vartheta) ]
\end{split}
\end{equation}
From Equation \ref{policy_gradient} we have that the gradient of $J$ w.r.t. $\vartheta$ is equal to the expected value of the function $G(\tau, \vartheta) = R(\tau) \triangledown_{\vartheta} \log p(\tau; \vartheta)$. This function provides an unbiased estimate of the gradient of $J$ and can be computed by running policy $\pi_\vartheta$ and sampling a trajectory $\tau$ without knowing the dynamics of the system since $p(x_{t+1}|x_t, u_t)$ does not depend on parameter $\vartheta$. Following this direction is equivalent to running stochastic gradient descent on $J$.

\begin{equation}
\triangledown_{\vartheta} \log p(\tau; \vartheta) = \sum_{t=0}^{L-1} \triangledown_{\vartheta} \log \pi_{\vartheta}(u_t|\tau_t)
\end{equation}
The policy gradient algorithm can be summarized:
\begin{enumerate}
\item Choose $\vartheta_0$, stepsize sequence $\alpha_k$, and set $k=0$; 
\item Run the simulator with policy $\pi_{\vartheta_k}$ and sample $\tau_k$;
\item $\vartheta_{k+1} = \vartheta_k + \alpha_k R(\tau_k) \sum_{t=0}^{L-1} \triangledown_{\vartheta} \log \pi_{\vartheta}(u_{tk}|\tau_t)$;
\item $k = k + 1$, go to step 2.
\end{enumerate}

The policy gradient algorithm can be run on any problem if sampling from $\pi_{\vartheta}$ can be done efficiently. Policy gradient is simple as it optimizes over a parametric family $p(u; \vartheta)$ instead of optimizing over the space of all probability distributions. However, there are constraints regarding the probability distribution, which should be easy to sample from, easy to search by gradient methods, and rich enough to approximate delta functions. In addition, the complexity of the method depends on the dimensionality of the search space and can be slow to converge. Finally, the policy gradient update is noisy, and its variance increases proportionally with the simulation length $L$.


The other solution to the problem of dealing with non-differentiable output is to use the the \textit{Gumbel-Softmax} \cite{jang2016categorical} approach, and replace the non-differentiable sample from the categorical distribution with a differentiable sample from a Gumbel-Softmax distribution. The Gumbel-Softmax distribution is a continuous distribution on the simplex that can approximate categorical samples. Parameter gradients can be easily computed by applying the reparameterization trick \cite{kingma2013auto}, a popular technique used in variational inference and adversarial learning of generative models in which the expectation of a measurable function $g$ of a random variable $\epsilon$ is calculated by integrating $g(\epsilon)$ with respect to the distribution of $\epsilon$:

\begin{equation}
\mathbb{E}(g(\epsilon)) = \int g(\epsilon) dF_{\epsilon}
\end{equation}
Therefore, in order to compute the expectation of $z=g(\epsilon)$ we do not need to know explicitly the distribution of $z$, but only know $g$ and the distribution of $\epsilon$. This can alternatively be expressed as:

\begin{equation}
\mathbb{E}_{\epsilon \sim p(\epsilon)}(g(\epsilon)) = \mathbb{E}_{z \sim p(z)}(z)
\end{equation}
If the distribution of variable $z$ depends on parameter $\phi$, i.e. $z \sim p_{\phi}(z)$, and if we can assume $z=g(\epsilon, \phi)$ for a known function $g$ of parameters $\phi$ and noise distribution $\epsilon \sim \mathcal{N} (0,1)$, then for any measurable function $f$:

\begin{equation}
\label{reparameterization}
\begin{split}
\mathbb{E}_{\epsilon \sim p(\epsilon)}(f(g(\epsilon, \phi))) &= \mathbb{E}_{z \sim p_{\phi}(z)}(f(z)) \\
\mathbb{E}_{\epsilon \sim p(\epsilon)}(\triangledown f(g(\epsilon, \phi))) &= \triangledown_{\phi} \mathbb{E}_{\epsilon \sim p(\epsilon)}(f(g(\epsilon, \phi))) \\  
&=  \triangledown_{\phi} \mathbb{E}_{z \sim p_{\phi}(z)}(f(z))
\end{split}
\end{equation}
In equation \ref{reparameterization}, $z$ has been conveniently expressed such that functions of $z$ can be defined as integrals w.r.t. to a density that does not depend on the parameter $\phi$. Constructing unbiased estimates of the gradient is done using Monte Carlo methods:
\begin{equation}
\triangledown_{\phi} \mathbb{E}_{z \sim p_{\phi}(z)}(f(z)) \sim \frac{1}{M}\sum_{i=1}^{M}\triangledown f(g(\epsilon^{i}, \phi))
\end{equation}
The reparameterization trick aims to make the randomness of a model an input to that model instead of letting it happen inside the model. Given this, the network model is deterministic and we can differentiate with respect to sampling from the model. An example of applying the reparameterization trick is to rewrite samples drawn from the normal distribution $z \sim \mathcal{N}(\mu, \sigma)$ as $z=\mu + \sigma \epsilon $, with $\epsilon \sim \mathcal{N}(0,1)$. In this way stochastic nodes are avoided during backpropagation. However, the re-parameterization trick cannot be directly applied to discrete valued random variables, for eg. text data, as gradients cannot backpropagate through discrete nodes in the computational graph.

The Gumbel-Softmax trick attempts to overcome the inability to apply the reparameterization trick to discrete data. It parameterizes a discrete distribution in terms of a Gumbel distribution, i.e. even if the corresponding function is not continuous, it will be made continuous by applying a continuous approximation to it. A random variable $G$ has a standard Gumbel distribution if $G=-\log(-\log(U)), U \sim \text{Unif}[0,1]$. Any discrete distribution can be parameterized in terms of Gumbel random variables as follows. If $X$ is a discrete random variable with $P(X=k) \propto \alpha_k$ random variable and $\{G_{k}\}_{k \le K}$ an i.i.d. sequence of standard Gumbel random variables, then:

\begin{equation}
\label{sampling_gumbel}
X = \arg \max_k(\log \alpha_k +G_k)
\end{equation} 
Equation \ref{sampling_gumbel} illustrates sampling from a categorical distribution: draw Gumbel noise by transforming uniform samples, add it to $\log \alpha_k$, then take the value of $k$ that yields the maximum. The $\arg \max$ operation that relates the Gumbel samples is not continuous, however discrete random variables can be expressed as one-hot vectors and take values in the probability simplex:


\begin{equation}
\Delta^{K-1}= \{ x  \in R^{K}_{+}, \sum_{k=1}^{K}x_k=1 \}
\end{equation}
A $\text{one\_hot}$ vector corresponds to a discrete category, and since the $\arg \max$ function is not differentiable, a softmax function can be used instead as a continuous approximation of $\arg \max$:

\begin{equation}
f_{\tau}(x)_k = \frac{\exp(x_k/\tau)}{\sum_{k=1}^{K}\exp(x_k/ \tau)}
\end{equation}
Therefore, the sequence of simplex-valued random variables $X^{\tau}$ is: 





\begin{equation}
\begin{split}
\label{eq_GumbelSoftmax_Distribution}
X^{\tau} = (X_k^{\tau})_k &= f_{\tau}(\log \alpha + G) \\
&= \frac{\exp((\log \alpha_k + G_k)/ \tau)}{\sum_{i=1}^{K}\exp((\log \alpha_i + G_i)/\tau)}
\end{split}
\end{equation} 
Equation \ref{eq_GumbelSoftmax_Distribution} is known as the Gumbel-Softmax distribution and can be evaluated exactly for different values of $x$, $\alpha$ and $\tau$, where $\tau$ is a temperature parameter that controls how closely the samples from the Gumbel-Softmax distribution approximate those from the categorical distribution. When $\tau \rightarrow 0 $, the softmax function becomes an $\arg \max$ function and the Gumbel-Softmax distribution becomes the categorical distribution. At training time $\tau$ is a set to a value greater than 0 which allows gradients to backpropagate past the sample, and then is gradually annealed to a value close to 0. The Gumbel Softmax trick is important as it allows for the inference and generation of discrete objects. A direct application of this technique is generating text via GANs.

In summary, GANs have shown impressive performance at generating natural images nearly indistinguishable from real images, however applying GANs to text generation is a non-trivial task due to the special nature of the linguistic representation. According to Dai et al \cite{dai2017towards}, the two main challenges to overcome when using GANs with textual input are: 

\textit{i)} first, text generation is a sequential non-differentiable sampling procedure which samples a discrete token at each time step (vs. image generation where the transformation from the input random vector to the produced output image is a deterministic continuous mapping); the non-differentiability of text makes it difficult to apply back-propagation directly, and to this end, classical reinforcement learning methods such as Policy Gradient \cite{sutton2000policy} have been used. In  policy gradient the production of each word is considered as an action for which the reward comes from the evaluator, and gradients can be back-propagated by approximating the stochastic policy with a parametric function. 

\textit{ii)} second, in the GAN setting the generator receives feedback from the evaluator when the entire sample is produced, however for sequence generation this causes difficulties during training, such as vanishing gradients and error propagation. To allow the generator to get early feedback when a text sequence is partly generated, Monte Carlo rollouts are used to calculate the approximated expected future reward. This has been found empirically to improve the efficiency and stability of the training process.

Unlike in conventional GAN settings that deal with image generation, the production of sentences is a discrete sampling process, which is also non-differentiable. A natural question that arises is how can the feedback be back-propagated from the discriminator to the generator under such a formulation. Policy gradient considers a sentence as a sequence of actions, where each word $w_t$ is an action and the choices of such actions are governed by a policy $\pi_{\theta}$. The generative procedure begins with an initial state $S_{1:0}$ which is the empty sentence, and at each time step $t$ the policy $\pi_{\theta}$ takes as input the previously generated words  $S_{1:t-1}$ up until time $t-1$, as well as the noise vector $z$, and yields a conditional distribution $\pi_{\theta}(w_t | z, S_{1:t-1})$ over the vocabulary words. The computation is done one step at a time moving along the LSTM network and sampling an action $w_t$ from the conditional distribution up until $w_t$ will be equal to the end of sentence indicator, in which case the sentence is terminated. The reward for the generated sequence of actions $S$ will be a score $r$ calculated by the discriminator. However, this score can be computed only after the sentence has been completely generated, and in practice this leads to difficulties such as vanishing gradients and very slow training convergence. Early feedback is used to evaluate the expected future reward when the sentence is partially generated, and the expectation can be approximated using Monte Carlo rollouts. The Monte Carlo rollout method is suitable to use when  a part of the sentence $S_{1:t}$ has been already generated, and we continue to sample the remaining words of the sentence from the LSTM network until the end of sentence token is encountered. The conditional simulation is conducted $n$ times, which results in $n$ sentences. For each sentence we compute an evaluation score, and the rewards obtained by the simulated sentences are averaged to approximate the expected future reward of the current sentence. In this way updating the generator is possible with feedback coming from the discriminator. The utility of the policy gradient method is that by using the expected future reward the generator is provided with early feedback and becomes trainable with gradient descent. 

Yu et al propose SeqGAN \cite{yu2017seqgan}, a GAN-based sequence generation framework with policy gradient, which is the first work to employ GANs for generating sequences of discrete tokens to overcome the limitations of GANs on textual data. SeqGAN treats the sequence generation procedure as a sequential decision making process \cite{bachman2015data}. A discriminator is used to evaluate the generated sequence and provide feedback to the generative model to guide its learning. It is a well known problem of GANs that for text data (discrete ouputs) the gradient cannot be passed back from the discriminator to the generator. SeqGAN addresses this problem  by treating the generator as a stochastic parameterized policy trained via policy gradient \cite{sutton2000policy} and optimized by directly performing gradient policy update, therefore avoiding the differentiation difficulty for discrete data. The reinforcement learning reward comes from the discriminator based on the likelihood that it would be fooled judged on a complete sequence of tokens, and is passed back to the intermediate state-action steps using Monte Carlo search \cite{browne2012survey}. 

The sequence generation problem is defined as follows. Given a dataset of human written sequences, train a generative model $G_{\theta}$ parameterized by $\theta$ to output sequence $Y_{1:T} = (y_1, \ldots, y_t, \ldots, y_T), y_t \in Y$, where $Y$ is the word vocabulary. The current state is the sequence of tokens $(y_1, \ldots, y_{t-1})$ generated until timestep $t$, and the action $a$ taken from this state is the selection of next token $y_t$. The policy model $G_{\theta}(y_t|Y_{1:t-1})$ is stochastic and will select an action according to the leant probability distribution of the input tokens. The state transition from the current state $s=Y_{1:t-1}$ to the next state $s^{'} = Y_{1:t}$ after choosing action $a=y$ is deterministic, i. e. $\delta_{s,s^{'}}^{a}=1$ for next state $s^{'}$, and $\delta_{s,s^{''}}^{a}=0$ for other next states $s^{''}$. The discriminative model $D_{\phi}(Y_{1:T})$ is used to guide the generator $G_{\theta}$, and outputs a probability indicating how likely a sequence $Y_{1:T}$ produced by $G_{\theta}$ comes from real sequence data. $D_\phi$ is trained with both real and fake examples from the real sequence data and the synthetic data generated by $G_{\theta}$. The objective of the generator model (policy) $G_{\theta}(y_y| Y_{1:t-1})$ is to maximize its expected end reward  $R_T$ which comes from the discriminator $D_{\phi}$ for a sequence which is generated starting from initial state $s_0$: 

\begin{equation}
J(\theta) = \mathbb{E}[R_T|s_0, \theta] = \sum_{y_1 \in Y}G_{\theta}(y_1|s_0)Q_{D_{\phi}}^{G_{\theta}}(s_0, y_1)
\end{equation}
The action-value function $Q_{D_{\phi}}^{G_{\theta}}(s, a)$ for a sequence represents the expected cumulative reward starting from state $s$, taking action $a$ and then following policy $G_{\theta}$. The action value function $Q_{D_{\phi}}^{G_{\theta}}(s, a)$ is calculated as the estimated probability (reward) the discriminator $D_{\phi}(Y_{1:T}^{n})$ assigns to the generated sample being real:

\begin{equation}
Q_{D_{\phi}}^{G_{\theta}}(a=y_T, s=Y_{1:T-1}) = D_{\phi}(Y_{1:T}^{n})
\end{equation}
In the GAN setup, the discriminator $D_{\phi}$ can only provide a reward at the end of a finished sequence. In order to evaluate the action-value function $Q_{D_{\phi}}^{G_{\theta}}(s, a)$ for an intermediate state $s$, Monte Carlo search with roll-out policy $G_{\beta}$ (identical to the generator $G_{\theta}$ policy) is used to sample the unknown remaining $T-t$ tokens that result in a complete sentence. The roll-out policy  $G_{\beta}$ starts from the current state $s$ and is run for $N$ times to get an accurate assessment of the action-value function $Q_{D_{\phi}}^{G_{\theta}}(s, a)$ through a batch of $N$ output samples, thus reducing the variance of the estimation: 

\begin{equation}
\begin{split}
\{Y_{1:T}^{1}, \ldots, Y_{1:T}^{N}\}&=MC^{G_{\beta}}(Y_{1:t}; N) \\
Q_{D_{\phi}}^{G_{\theta}}(a=y_t, s=Y_{1:t-1})
&=\begin{cases}
\frac{1}{N} \sum_{n=1}^{N}D_{\phi}(Y_{1:T}^n), \\ \text{if } Y_{1:T}^n \in MC^{G_{\beta}}(Y_{1:t; N}), t < T \\
D_{\phi}(Y_{1:t}), \text{if } t = T
\end{cases}
\end{split}
\end{equation}
The generator starts with random sampling at first, but once more realistic samples have been generated, the discriminator $D_{\phi}$ is updated (which will in turn improve the generator model iteratively):

\begin{equation}
\min_{\phi}-\mathbb{E}_{Y \sim p_{\text{data}}}[\log D_{\phi}(Y)] - \mathbb{E}_{Y \sim G_{\theta}}[\log (1 - D_{\phi}(Y))] 
\end{equation}
The generator $G_{\theta}$ is updated every time a new discriminator $D_{\phi}$ has been obtained. The gradient of the generator's objective function $J(\theta)$ w.r.t the generator's parameters $\theta$ is expressed as follows:

\begin{equation}
\begin{split}
\nabla_{\theta} J(\theta) = \sum_{t=1}^{T}\mathbb{E}_{Y_{1:t-1} \sim G_{\theta}}\bigg[\sum_{y_t \in Y} \nabla_{\theta}G_{\theta}(y_t|Y_{1:t-1}) \cdot \\
\cdot Q_{D_{\phi}}^{G_{\theta}}(Y_{1:t-1}, y_{t})\bigg]
\end{split}
\end{equation}
Expectation $\mathbb{E}$ can be approximated by sampling methods, and generator's parameters are updated:

\begin{equation}
\theta \leftarrow \theta + \alpha_{h}\nabla_{\theta}J(\theta), \text{where } \alpha_{h} - \text{learning rate}
\end{equation}
In the initial stages of training, the generator $G_{\theta}$ is pre-trained via maximum likelihood estimation, and the discriminator $D_{\phi}$ is pre-trained via minimizing the cross-entropy between the ground truth label and the predicted probability; after the pre-training stage is over, the generator and the discriminator are trained alternatively. The SeqGAN authors chose an LSTM \cite{schmidhuber1997long} architecture for the generator in order to avoid the vanishing and the exploding gradient problem of back-propagation through time, and a CNN \cite{lecun1998gradient}, \cite{kim2014convolutional} architecture with highway networks \cite{srivastava2015highway} as discriminator. The evaluation metric is set to minimize the average negative log-likelihood between the generated data and an oracle considered as the human observer:

\begin{equation}
\text{NLL}_{\text{oracle}} = -\mathbb{E}_{Y_{1:T \sim G_{\theta}}}\bigg[\sum_{t=1}^{T} \log G_{\text{oracle}}(y_t|Y_{1:t-1}) \bigg]
\end{equation}
Lin et al \cite{lin2017adversarial} consider that GANs restrict the discriminator too much by forcing it to be a binary classifier. Because of this setup, the discriminator is limited in its learning capacity especially for tasks with a rich structure, such as when generating natural language expressions. The authors propose a generative adversarial framework called RankGAN, which is able to capture the richness and diversity of language by learning a relative ranking model between the machine written and human written sentences in an adversarial framework. The adversarial network consists of two neural network models, a generator $G_{\theta}$ and a ranker $R_{\phi}$, where $\theta$ and $\phi$ are parameters. The RankGAN discriminator $R_{\phi}$, instead of performing a binary classification task as in conventional GANs, is trained to rank the machine-written sentences lower than human-written sentences w.r.t. a human-written reference set. Alternatively, the generator $G_{\theta}$ is trained to confuse the ranker $R$ in such a way that machine written sentences are ranked higher than human written sentences with regard to the reference set. The authors consider that by viewing a set of samples collectively (instead of just one sample) and evaluating their quality through relative ranking, the discriminator can make better judgements regarding the quality of the samples, which helps in turn the generator better learn to generate realistic sequences. The problem can be expressed mathematically as $G_{\theta}$ and $R_{\phi}$ playing a minimax game with the objective function $\mathcal{L}$:

\begin{equation}
\label{RankGANobj}
\begin{split}
\min_{\theta}\max_{\phi} \mathcal{L}(G_{\theta}, R_{\phi}) = \mathbb{E}_{s \sim P_h}[\log R_{\phi}(s|U, C^{-})] + \\ 
\mathbb{E}_{s \sim G_{\theta}}[\log(1-R_{\phi}(s|U, C^{+}))]
\end{split}
\end{equation}
The ranker $R_{\phi}$ is optimized to increase the likelihood of assigning a high probability to the real sentence $s$ and a low probability to the fake generated data $G_{\theta}$. $s \sim P_h$ denotes that sentence $s$ is sampled from human written sentences, while $s \sim G_{\theta}$ denotes that sentence $s$ is sampled from machine written sentences. $U$ is a reference set which is used for estimating  relative ranks. $C^{+}$ and $C^{-}$ are comparison sets with regards to input sentences. When the input sentence $s$ is sampled from the real data, $C^{-}$ is sampled from the generated data, and alternatively when the sentence $s$ is sampled from the synthetic data generated by $G_{\theta}$, $C^{+}$ is sampled from human written data.

Similar to SeqGAN, the authors use policy gradient to overcome the non-differentiability problem of text data.  However, unlike SeqGAN, the regression based discriminator is replaced with a ranker and a new learning objective function. The generative model $G_{\theta}$ is an LSTM network, while the ranker $R_{\phi}$ is a CNN network. The rewards for training the model are encoded with relative ranking information. When a sequence is incomplete, an intermediate reward is computed using Monte Carlo rollout methods. The expected future reward $V$ for partial sequences is defined as:

\begin{equation}
\label{rankgan_reward}
V_{\theta, \phi}(s_{1:t-1, U}) = \mathbb{E}_{s_r \sim G_{\theta}}[ \log R_{\phi}(s_r | U, C^{+}, s_{1:t-1}) ]
\end{equation}
In Equation \ref{rankgan_reward} above, $s_r$ denotes a complete sequence sampled by using rollout methods starting from sequence $s_{1:t-1}$. A total of $n$ different paths are sampled, and their corresponding ranking scores are computed. The average ranking score is used to approximate the expected future reward for the current partially generated sequence $s_{1:t-1}$; the ranking score of an input sentence $s$ given reference sentence $u$ and comparison set $C$ (where $C=C^{+}$ if sentence $s$ is machine generated, $C=C^{-}$ otherwise) is computed using a softmax-like formula:

\begin{equation}
\label{rankgan_ranking}
\begin{split}
P(s|u,C) &=\frac{\exp(\gamma \alpha(s|u))}{\sum_{s^{'} \in C^{'}} \exp(\gamma \alpha (s^{'}|u))}, \text{where }\\
\alpha(s|u) &=\cos(y_s,y_u)=\frac{y_s  y_u}{||y_s||  ||y_u||}
\end{split}
\end{equation}
In Equation \ref{rankgan_ranking}, $y_s$ is the embedded feature vector of the input sentence, and $y_u$ is the embedded feature vector of the reference sentence. The gradient of the objective function for generator $G_{\theta}$ for start state $s_0$, vocabulary $V$, and generator policy $\pi_\theta$ is computed as:

\begin{equation}
\begin{split}
\triangledown_{\theta} \mathcal{L}_{\theta}(s_0) = \mathbb{E}_{s_{1:T}\sim G_{\theta}} \bigg[\sum_{t=1}^{T}\sum_{w_t \in V} \triangledown_{\theta} \pi_{\theta} (w_t | s_{1:t-1}) \cdot \\
\cdot V_{\theta, \phi} (s_{1:t}, U)\bigg]
\end{split}
\end{equation}
Therefore, RankGAN deals with the gradient vanishing problem of GAN by replacing the original binary classifier discriminator with a ranking model in a learning-to-rank framework. The ranking score is computed by taking a softmax over the expected cosine distances from the generated sequences to the real data.

Guo et al \cite{guo2017long} find that a limitation of current GAN frameworks for text generation \cite{yu2017seqgan}, \cite{lin2017adversarial}, \cite{rajeswar2017adversarial}, \cite{che2017maximum}, \cite{li2017adversarial}, \cite{zhang2017adversarial} is that they are only capable of generating short texts, within a limited length of around 20 tokens. Generating longer sequences is a less studied but more challenging research problem with a lot of useful applications, such as the auto-generation of news articles or product descriptions. Nevertheless, long text generation faces the issue that the binary guiding signal from generator $D$ is sparse and non-informative; it does not provide useful information regarding the intermediate syntactic structure and semantics of the generated text so that the generator $G$ could learn from that signal. Besides that, it is only available after the entire sequence has been generated, and the final reward value does not provide much guidance on how to alter the parameters of $G$ at training time. Moreover, the approach of relying on binary feedback from the discriminator requires a very large number of real and generated samples to improve $G$. Aiming to make the guiding signal coming from the discriminator $D$ more informative, the authors propose LeakGAN \cite{guo2017long}, a GAN approach for adversarial text generation in which the discriminative model $D$ is allowed to leak its own high-level extracted features (in addition to providing the final reward value) to better guide the training of the generative model $G$. The authors pick a hierarchical generator for $G$, which is made up of two distinct modules: a \textit{high-level manager} module, and a \textit{low-level worker} module. The high level manager module (or mediator) receives the feature map representation of the discriminator $D$; this is not normally allowed in the conventional GAN setup as this feature map is internally maintained by the discriminator. The manager embeds this feature map representation coming from the discriminator and passes it over to the worker module. The worker first encodes the current generated sequence, and combines this resulting encoding with the embedding produced by the manager to decide what action to take at the current state. Therefore, LeakGAN ``leaks'' guiding signals from the discriminator $D$ to the generator $G$ more frequently and more informatively throughout the sequence generation process and not at the end only, helping $G$ improve better and faster. 


The discriminator $D_{\phi}$ is made up of a feature extractor $\mathcal{F}(.; \phi_f)$ and a final sigmoid classification layer. For input sequence $s$, $D_{\phi}$ is defined as:

\begin{equation}
D_{\phi}(s) = \text{sigmoid}(\phi_l^T \mathcal{F} ( s; \phi_f)) = \text{sigmoid}(\phi_l^T f)
\end{equation}
The feature vector in the last layer of $D_\phi$ is denoted as $f = \mathcal{F} ( s; \phi_f) $, and it will be leaked to the generator $G_\theta$. A natural implication of this approach is that the reward the generator $G_\theta$ receives for a partially generated sequence is directly related to the quality of the extracted features by the discriminator $D_{\phi}$. Therefore, for the discriminator $D_{\phi}$ to yield a high reward, it is necessary to find a highly rewarding region in the extracted feature space. The authors consider that compared to a scalar signal, the feature vector $f$ is more informative as it captures the position of the generated words in the extracted feature space. $D_{\phi}$ is implemented as a CNN network.  The manager module $\mathcal{M}(f_t, h_{t-1}^{M}; \theta_m)$ of the hierarchical generator $G_\theta$ receives as input the extracted feature vector $f_t$, which it combines with its internal hidden state to produce the goal vector $g_t$:

\begin{equation}
\begin{split}
g_t^{'} &= \mathcal{M}(f_t, h_{t-1}^{M}; \theta_m)\\
g_t &= \frac{g_t^{'}}{||g_t^{'}||}
\end{split}
\end{equation}
The goal vector embedding $w_t$ of goal $g_t$ is computed by applying a linear transformation $\psi$ with weight matrix $W_{\psi}$ to the sum of recent $c$ goals:

\begin{equation}
w_t = \psi (\sum_{i=1}^{c}g_{t-i}) = W_{\psi} (\sum_{i=1}^{c}g_{t-i})
\end{equation}
$w_t$ is fed to the worker module $\mathcal{W}(.;\theta_w)$, which is in charge with the generation of the next token. The worker module takes the current word $x_t$ as input and outputs matrix $O_t$; this matrix is then combined through a softmax with the goal vector embedding $w_t$:

\begin{equation}
\begin{split}
O_t, h_t^W &= \mathcal{W}(x_t, h_{t-1}^{W}; \theta_w) \\
G_{\theta}(.|s_t) &= \text{softmax}(O_t w_t / \alpha)
\end{split}
\end{equation}
At training time, the manager and the worker modules are trained separately -- the manager is trained to predict which are the most rewarding positions  in the discriminative feature space, while the worker is rewarded to follow these directions. The gradient for the manager module is defined as:

\begin{equation}
\triangledown _{\theta_m}^{\text{adv}}g_t = -Q_{\mathcal{F}}(s_t, g_t)\triangledown_{\theta_m} d_{\cos}(f_{t+c}-f_t, g_t(\theta_m))
\end{equation}
$Q_{\mathcal{F}}(s_t, g_t)$ defines the expected reward under the current policy and can be approximated using Monte Carlo search. $d_{\cos}$ computes cosine similarity between the goal vector $g_t(\theta_m)$ produced by the manager, and the change in feature representation $f_{t+c} -f_t$ after $c$ transitions. In order to achieve a high reward, the loss function is trying to force the goal vector to match the transition in feature space. Before the adversarial training takes place, the manager undergoes a pre-training stage with  a separate training scheme which mimics the transition of real text samples in the feature space:

\begin{equation}
\triangledown_{\theta_m}^{\text{pre}} = -\triangledown_{\theta_m}d_{\cos}(f_{t+c}^{'} - f_t^{'}, g_t(\theta_m))
\end{equation}
The worker uses the REINFORCE algorithm during training to maximize the reward when taking action $x_t$ given the previous state is $s_{t-1}$:

\begin{equation}
\begin{split}
\triangledown_{\theta_w}\mathbb{E}_{s_{t-1} \sim G} \bigg[ \sum_{x_t} r_{t}^{I} \mathcal{W}(x_t|s_{t-1}; \theta_w)\bigg] &= \\ \mathbb{E}_{s_{t-1} \sim G, x_t \sim \mathcal{W} (x_t | s_{t-1})}\bigg[ r_t^{I} \triangledown_{\theta_w} \log \mathcal{W} (x_t|s_{t-1}; \theta_w) \bigg]  \\
r_t^{I} = \frac{1}{c} \sum_{i=1}^{c}d_{\cos}(f_t - f_{t-i}, g_{t-i})
\end{split}
\end{equation}
 During the adversarial training process, the generator $G_\theta$ and the discriminator $D_\phi$ are trained in alternative stages. When the generator $G_\theta$ is trained, the worker $\mathcal{W}(.;\theta_w)$ and the manager $\mathcal{M}(.;\theta_m)$ modules are trained alternatively fixing each other.

Mode collapse \cite{goodfellow2016nips} is a common problem when training GAN models, when the generator learns to produce samples with extremely low variety, limiting the usefulness of the leant GAN model. In mode collapse the generator network learns to output samples from a few modes of the data distribution only, missing out on many other modes even though samples from these missing modes can be found throughout the training data. Mode collapse can range from complete collapse, when the generated samples are entirely identical, to partial collapse when the generated samples present some  common properties \cite{srivastava2017veegan}, \cite{salimans2016improved}. Several attempts have been made to address the problem, which include: \textit{i)} directly encouraging the generator cost function to account for the diversity of the generated batches by comparing these samples across a batch in order to determine whether the entire batch is real or fake, \textit{ii)} anticipate counterplay, in which the generator learns to fool the discriminator before the discriminator has a chance to respond (and therefore taking counterplay into account), \textit{iii)} experience replay, which minimizes the switching between modes by showing old fake generated samples to the discriminator every now and then, and \textit{iv)} using multiple GANs, in which a GAN is trained for each different mode so that when combined, the GANs altogether cover all modes.

In LeakGAN, in order to address mode collapse, the authors propose an interleaved training scheme, which combines supervised training using maximum likelihood estimation with GAN adversarial training (instead of carrying only GAN adversarial training after the pretraining stage). Blending two training schemes is considered useful by the authors as it helps LeakGAN overcome local minimums, alleviates mode collapse and acts as an implicit regularizer on the generative model.

\subsection{Samples produced by the review generators}
\label{appendix_user_study_samples}

\begin{figure*}[!htbp]
\centering
  \includegraphics[width=\textwidth]{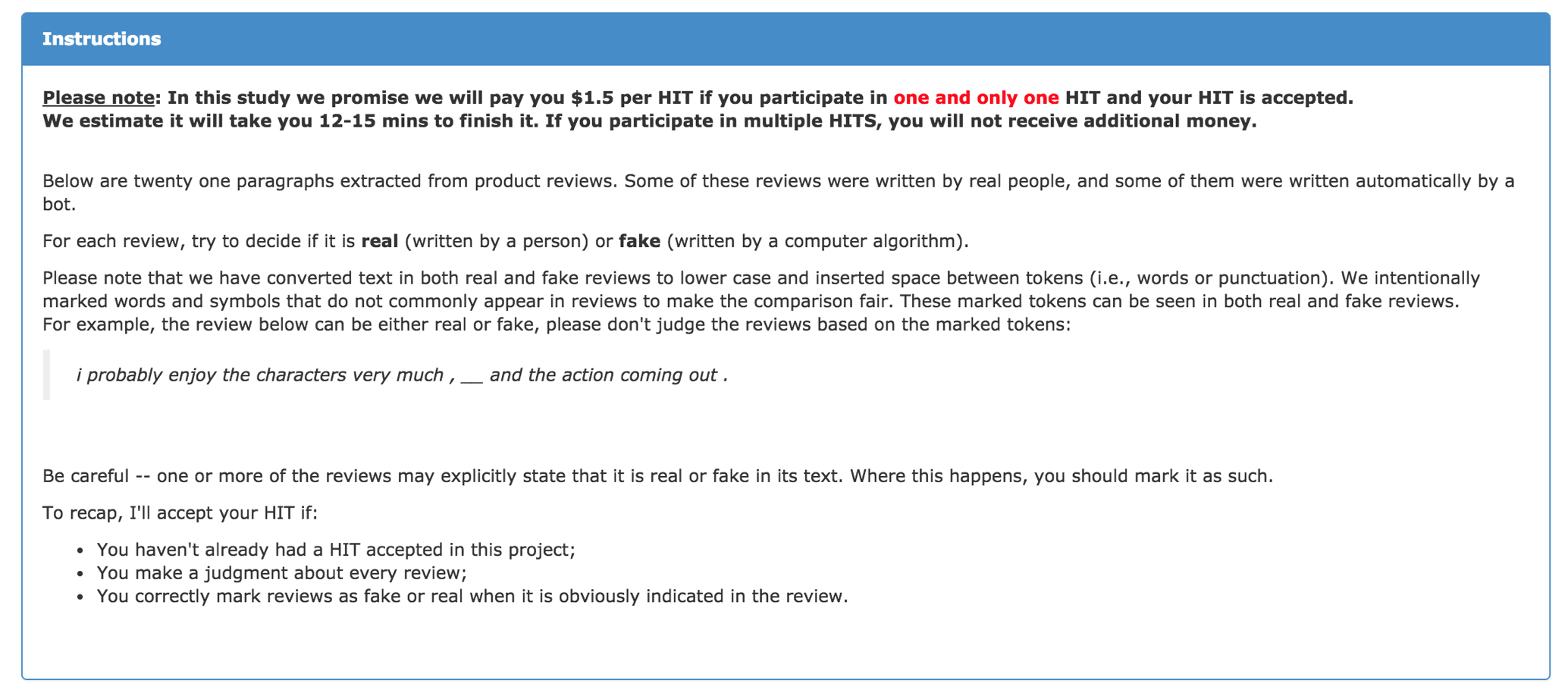}
  \caption{Screenshot of the instructions presented to Amazon Mechanical Turk workers.}
  \label{user_instructions}
\end{figure*}

\begin{figure*}[!htbp]
\centering
  \includegraphics[width=6.5in]{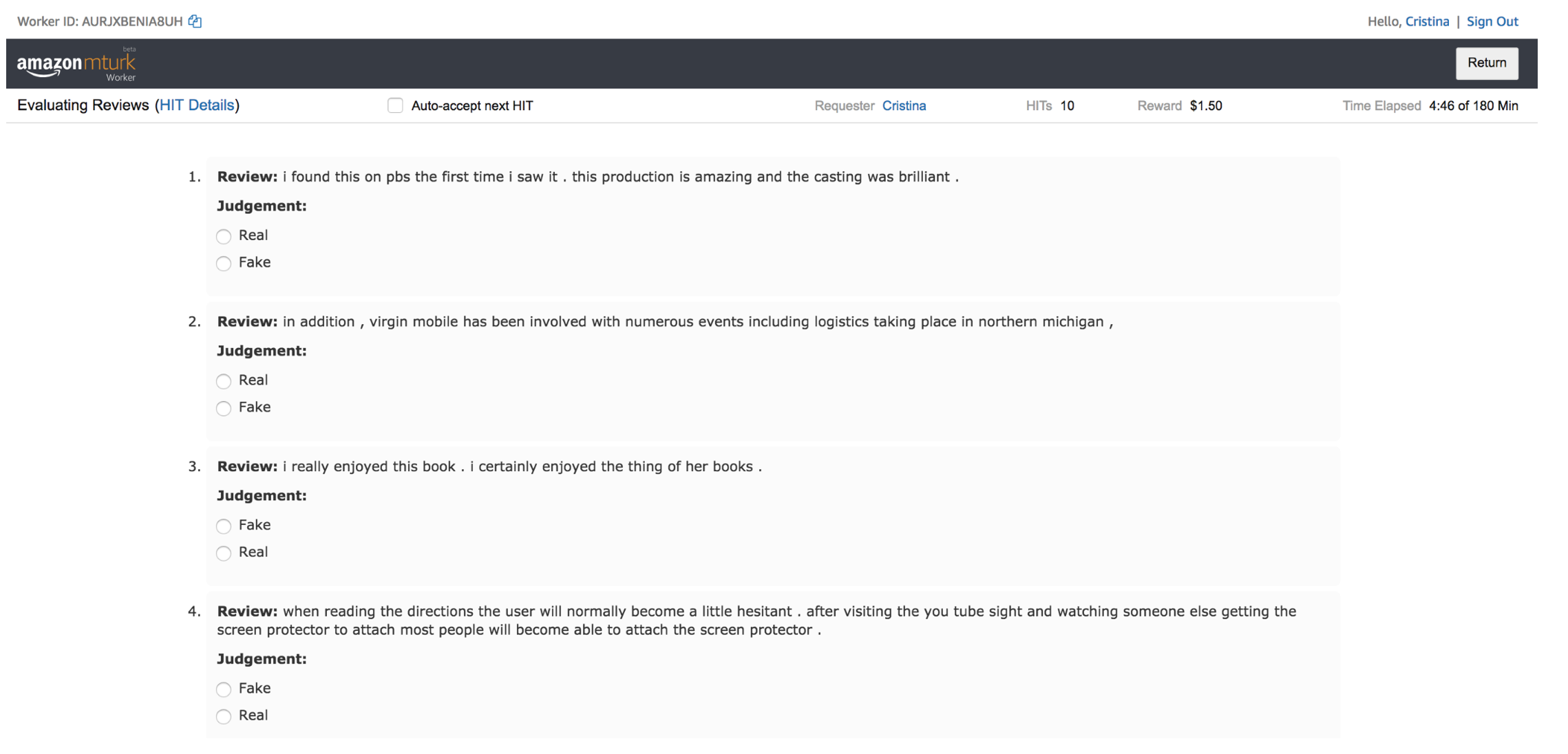}
  \caption{Screenshot of the Amazon Mechanical Turk user study interface.}
  \label{user_interface}
\end{figure*}

Figure \ref{user_instructions} shows the instructions given to the AMT workers who participated in this study. In Figure \ref{user_interface} we include a screen-shot of the user interface when annotating reviews.

In what follows we present samples generated by the review generators on which human annotators disagree most on whether these are human-written or machine-generated.
\begin{itemize}
  \item Word LSTM temp 1.0
  \begin{enumerate}[label=\alph*)]
  \item  i so enjoyed this book . i felt \underline{\hspace{0.4cm}} though . i especially like loving horses in the \underline{\hspace{0.4cm}} . and the story is well written .
  \item one of a different type on locked paranormal / vacation book . i enjoyed the characters and the plot . great mixture of historical fiction . 
  \item this first edition of the complete series 8 years over six episodes just makes you laugh . the original tv is by far my cup of tea !

  \item  works out of the box ! wouldn ' t spend the money for a better keyboard . use this with the matching kindle screen as well .
  
  \end{enumerate}
  
  \item Word LSTM temp 0.7
  \begin{enumerate}[label=\alph*)]
  \item i am looking forward to the next book . i am a \underline{\hspace{0.4cm}} \underline{\hspace{0.4cm}} and i enjoyed the story . i like books where the characters are real .
  \item this is an exciting book i could n ' t put down . i will probably read more books by this author . this is a must read .
  \item okay , that ' s how i expected this movie . it was okay but it was so boring . i was bored and was disappointed .
  \item this cable is not bad . it is so cheap and it works great . i ' ve used this for a couple of months now and \underline{\hspace{0.4cm}} on the ipad 
  \end{enumerate}
  
  \item Word LSTM temp 0.5
  
  \begin{enumerate}[label=\alph*)]
  
  \item  this book was a great read ! the story was exciting and a bit \underline{\hspace{0.4cm}} . i really enjoyed the characters and the story line .
  
  \item this is a great cable for the price . i would recommend this product to anyone needing a cable for a great price .
  
  \item this is a great series . it is a must see for anyone who loves period dramas . i love the \underline{\hspace{0.4cm}} .

  \item  these batteries seem to be working as expected . i have had no problems with this product . i would recommend this to anyone .
  
  \end{enumerate}
  
  \item Scheduled Sampling
  
  \begin{enumerate}[label=\alph*)]
  
  \item like most of the ones i have ! the tablet that came starts working properly .
  
  \item i have had any almost using keyboards with an iphone case and kept it nicely and time . and it works well .
  
  \item have got to watch it many times again and the seasons of \underline{\hspace{0.4cm}} each episode we can all watch it . 
  
  \item very interesting characters and likable characters that grow when you gave me \underline{\hspace{0.4cm}} of the \underline{\hspace{0.4cm}} because of the dog . what can i say is i absolutely loved it .
  
  \end{enumerate}
  
  \item Google LM
  
  \begin{enumerate}[label=\alph*)]
  
  \item  \underline{\hspace{0.4cm}} systems generally require less bandwidth and \underline{\hspace{0.4cm}} with operating systems , \underline{\hspace{0.4cm}} users to write and edit data nearly anywhere .
  
  \item seems all but impossible to access . \underline{\hspace{0.4cm}} is all a \underline{\hspace{0.4cm}} and gets a bad \underline{\hspace{0.4cm}} on every \underline{\hspace{0.4cm}} .
  
  \item \underline{\hspace{0.4cm}} is based in \underline{\hspace{0.4cm}} \underline{\hspace{0.4cm}} , \underline{\hspace{0.4cm}} , with a commercial office in \underline{\hspace{0.4cm}} 
  
  \item oved this clip and the \underline{\hspace{0.4cm}} and \underline{\hspace{0.4cm}} apps were about so much fun that \underline{\hspace{0.4cm}} paid a big price . \underline{\hspace{0.4cm}} 2 and 3 like crazy .
  
  \end{enumerate}
  
  \item Attention Attribute to Sequence
  
  \begin{enumerate}[label=\alph*)]
  
  \item i am always waiting for the next book to come out . i am a big fan of sean black and will .
  
  \item purchased this to use with my macbook pro . it worked out perfectly , as described . no complaints .
  
  \item great book all of the great mystery books . i enjoyed all of them and was sad when the book ended .
  
  \item  this is a great product . i ' ve had it for over a year now and it ' s still going strong . i ' m very happy with this purchase .
  
  \end{enumerate}
  
  \item Contexts to Sequences
  
  \begin{enumerate}[label=\alph*)]
  
  \item i love this series . i love the characters and the story . i love the characters and the story line .
  
  \item  a great book and a great read . i love the characters and the story . i would recommend this book to anyone .
  
  \item i enjoyed the story . it was a good read . i would recommend it to anyone who likes a good read .
  
  \item i love this book and i love the characters . i love this book and i was not disappointed . 

  \end{enumerate}
  
  \item Gated Contexts to Sequences
  
  \begin{enumerate}[label=\alph*)]
  
  \item this is the first book i have read by this author . would recommend to anyone who likes a good romance book . 
  
  \item one of the best books i have ever read . the chemistry between the two main characters was a good read .

  \item this book is awesome . lots of action and intrigue . i ' m glad i bought this book . thank you for sharing
  
  \item  great story and plot . sometimes a little slow at times but overall a good read .
  
  \end{enumerate}
  
  \item MLE SeqGAN
  
  \begin{enumerate}[label=\alph*)]
  
  \item you will like this movie - get this set \ldots better than expected award for the characters . bad ending . 
  \item this switch converter works fine with all games and works perfect , sturdy program to zero manual products . nice feel .
  
  \item i could not put it down . it was an interesting clean book , but i was expecting many more individuals in this story so i read in a long time . 
  
  \item great story . in college kids has been lost the \underline{\hspace{0.4cm}} mysteries , chris \underline{\hspace{0.4cm}} son is not better .
  
  \end{enumerate}
  
  \item SeqGAN
  
  \begin{enumerate}[label=\alph*)]
  
   \item it was slow he kept me interested , and i think i thoroughly enjoyed the story .
   
   \item i enjoyed this book and look forward to getting to \underline{\hspace{0.4cm}} larson .
   
   \item received in excellent condition . i thought it was great but didn ' t know that movies were more than high ratings which i am my cup of tea .
   
   \item awesome cute story . kudos to mr much \underline{\hspace{0.4cm}} of the sookie ' s story .
   
  \end{enumerate}
  
  \item RankGAN
  \begin{enumerate}[label=\alph*)]
  \item  robin williams is ok . just a great movie with  \underline{\hspace{0.4cm}} now . \underline{\hspace{0.4cm}} is a great film with three stars ! wonderful video for a very good movie .
  
  \item i have loved this movie so i could like the dvd sort of info . hot slow . love the old ford shows to though . \underline{\hspace{0.4cm}} a great actor .
  
  \item  this was a very amazing . \underline{\hspace{0.4cm}} laws and oh fact she became \underline{\hspace{0.4cm}} and \underline{\hspace{0.4cm}} is very unlikely together on the case .
  
  \item i say so i would that originally arrived so i love the circular inch screen . i am sad how it works .
  
  \end{enumerate}
  
  \item LeakGAN
  
  \begin{enumerate}[label=\alph*)]
      \item i really enjoyed reading this book . the author did an excellent job in delivering for all his writing books into us as business . a great summer read .
      
      \item just loved it , so much could read more of this series , i like it but it was not written in a book that is well written , but very interesting .
      
      \item i love hockey - baseball movie coming meets hockey ' s et addicted fear the birds feature so popular films have developed far worse reviews .
      
      \item a very good book with a lot of twists in this book . i will be checking out more of this author next book .
      
  \end{enumerate}

\end{itemize}


\subsection{Results}

\subsubsection{Human Evaluators}
\label{appendix_human_evaluators}

We chose the task of distinguishing machine-generated from real reviews because it is a straightforward surrogate of a Turing test. Moreover, how much their generated content can fool humans has been a key claim of many artificial intelligence models recently.  The low inter-rater agreement  suggests that this is a difficult task even for humans, which we hope would trigger the community to rethink about these claims. There are indeed finer-grained, perhaps more agreeable aspects of text quality (including semantic coherence, syntactic correctness, fluency, adequacy, diversity and readability). We decided not to include them in this experiment for two reasons: 1) as the first study, we are not sure which aspects human raters would consider when they judge for the realism of a review; 2) we wanted to keep the experiment design simple, and many of these aspects are harder to define. In the post-experiment survey, the raters commented on the reasons why they considered reviews as fake.

The low inter-rater agreement (0.27) reflects the difficulty/ subjectivity of the task: identifying individual reviews as human-written or machine-generated. Low human agreement is commonly reported in subjective evaluation tasks. Since our goal is to evaluate the \textbf{evaluators} instead of the competing algorithms, it is important to use a task neither too easy or too hard, so that there are distinguishable differences among the performances of competitors (including humans). 
When using the majority vote of human judgements, the accuracy of humans improved to a reasonable 72.63 \%.


\subsubsection{Discriminative Evaluators}
\label{appendix_discriminative_evaluators_results}

\begin{table}[!bp]
\caption{Accuracy of deep (LSTM) and shallow (SVM) meta-adversarial evaluators. \textbf{The lower the better.} Meta-adversarial evaluators do better than humans on individual reviews, with less bias between the two classes. 
GAN-based generators are considered to be the best by meta-adversarial evaluators.}  
\centering
\scalebox{0.8}{
\begin{tabular}{  l | c | c  }
\hline
\hline
\textbf{Generators} & \textbf{LSTM} & \textbf{SVM} \\
\hline
\hline
Word LSTM temp 1.0 & 48.29 \% &  \textbf{50.31} \%  \\
Word LSTM temp 0.7 & 92.58 \% &  78.69 \% \\
Word LSTM temp 0.5 & 99.31 \% & 94.74 \% \\
Scheduled Sampling & 50.09 \% & 51.31 \% \\
Google LM & 84.58 \% & 78.59 \% \\
Attention Attribute to Sequence & 90.08 \% & 74.37 \% \\
Contexts to Sequences & 100.00 \%  & 100.00 \% \\
Gated Contexts to Sequences & 98.37 \% & 96.26 \% \\
MLE SeqGAN & \textbf{41.45} \%  & 52.35 \% \\
SeqGAN & 50.05 \% & 56.20 \% \\
RankGAN & 66.28 \% & 70.17 \% \\
LeakGAN & 87.03 \% & 77.55 \% \\
\hline
D-test (all) & 77.58 \% & 74.50 \%  \\
D-test (human-written) & 80.12 \%  & 75.98 \% \\
D-test (machine-generated) & 75.04 \% & 73.01 \% \\
\hline
\hline
\end{tabular}}
\label{meta_discriminator_accuracy_rankings}
\end{table}

In Table \ref{meta_discriminator_accuracy_rankings} and Table \ref{meta_discriminator_accuracy_rankings_all} we present comprehensive results for the meta-adversarial evaluators.

\begin{table*}[!h]
\caption{Accuracy of deep (LSTM, CNN, CNN \& LSTM) and shallow (SVM, RF, NB, XGBoost) meta-adversarial evaluators. \textbf{The lower the better.} Meta-adversarial evaluators do better than humans on individual reviews, with less bias between the two classes. 
GAN-based generators are considered best by meta-adversarial evaluators.}  
\centering
\scalebox{0.7}{
\begin{tabular}{  l | c | c | c  | c | c | c | c  }
\hline
\hline
\textbf{Generators} & \textbf{LSTM} & \textbf{CNN} & \textbf{CNN \& LSTM} & \textbf{SVM} & \textbf{RF} & \textbf{NB } & \textbf{XGBoost} \\
\hline
\hline
Word LSTM temp 1.0 & 48.29 \% & 55.22 \% & 45.68 \% & \textbf{50.31} \% & 53.63 \% & 32.77 \% & 48.97 \% \\
Word LSTM temp 0.7 & 92.58 \% & 93.14 \% & 91.02 \% & 78.69 \% & 81.05 \% & 79.92 \% & 80.49 \%  \\
Word LSTM temp 0.5 & 99.31 \% & 99.35 \% & 99.08 \% & 94.74 \% & 94.29 \% & 96.86 \% & 94.71 \% \\
Scheduled Sampling & 50.09 \% & 48.77 \% & 43.37 \% & 51.31 \% & 52.88 \% & \textbf{20.97} \% & 44.12 \% \\
Google LM & 84.58 \% & 74.03 \% & 74.85 \% & 78.59 \% & 82.71 \% & 48.28 \% & 82.41 \% \\
Attention Attribute to Sequence & 90.08 \% & 91.78 \% & 89.94 \% & 74.37 \% & 77.29 \% & 80.02 \% & 71.68 \% \\
Contexts to Sequences & 100.00 \% & 100.00 \% & 99.97 \% & 100.00 \% & 99.98 \% & 100.00 \% & 99.98 \% \\
Gated Contexts to Sequences & 98.37 \% & 99.06 \% & 98.38 \% & 96.26 \% & 95.35 \% & 98.63 \% & 93.62 \%\\
MLE SeqGAN & \textbf{41.45} \% & \textbf{47.54} \% & \textbf{41.91} \%  & 52.35 \% & \textbf{51.14} \% & 21.83 \% & \textbf{43.71} \%\\
SeqGAN & 50.05 \% & 52.91 \% & 47.35 \% & 56.20 \% & 54.91 \% & 25.60 \% & 48.11 \% \\
RankGAN & 66.28 \% & 67.23 \% & 59.37 \% & 70.17 \% & 61.94 \% & 35.98 \% & 61.23 \% \\
LeakGAN & 87.03 \% & 80.28 \% & 79.57 \% & 77.55 \% & 67.74 \% & 46.80 \% & 63.80 \% \\
\hline
D-test (all) & 77.58 \% & 74.72 \% & 75.18 \% & 74.50 \% & 70.31 \% & 70.74 \% & 73.79 \%  \\
D-test (human-written) & 80.12 \% & 73.54 \% & 77.99 \% & 75.98 \% & 68.59 \% & 83.53 \% & 79.10 \%\\
D-test (machine-generated) & 75.04 \% & 75.90 \% & 72.38 \% & 73.01 \% & 72.04 \% & 57.95 \% & 68.48 \% \\
\hline
\hline
\end{tabular}}
\label{meta_discriminator_accuracy_rankings_all}
\end{table*}

\subsubsection{Text-Overlap Evaluators}
\label{appendix_text_overlap_evaluators_results}

\begin{figure*}[!htbp]
\centering
  \includegraphics[width=5in]{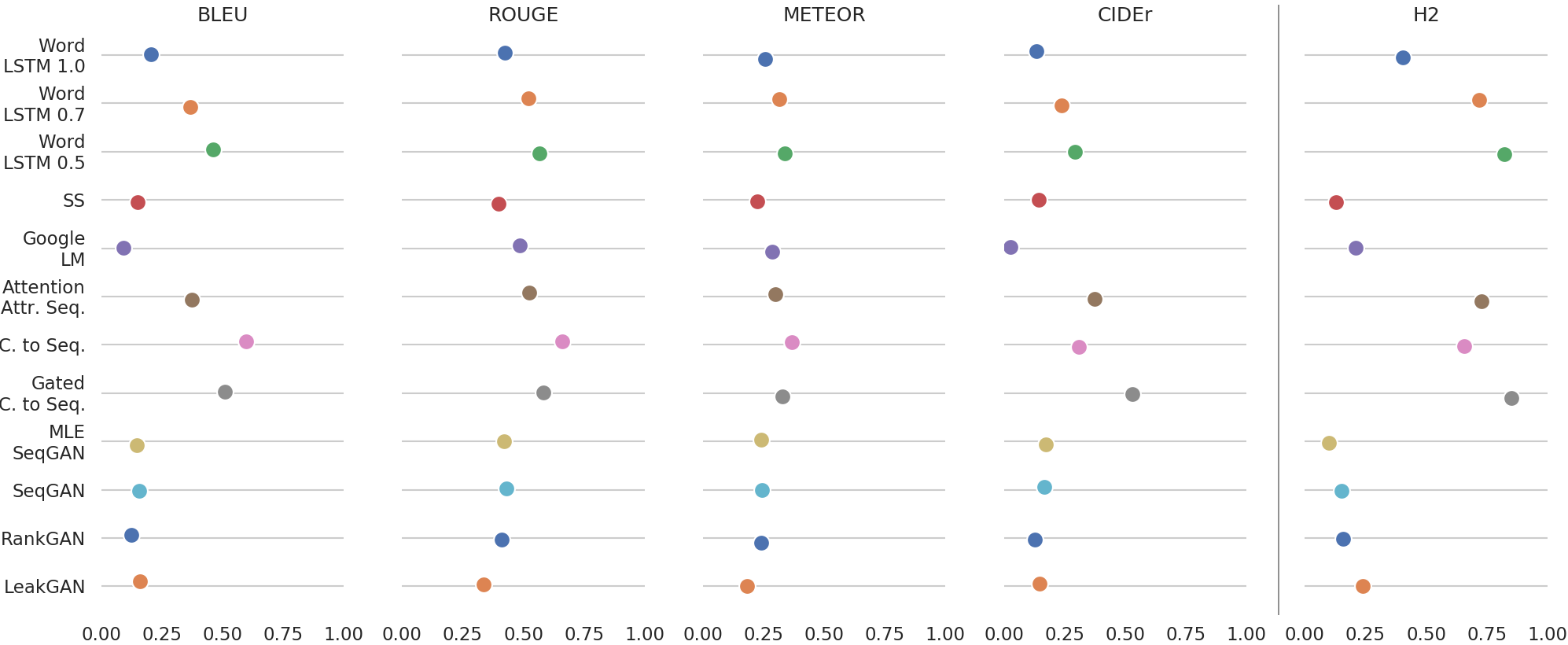}
  \caption{Text-Overlap Evaluators (BLEU, ROUGE, METEOR and CIDEr) scores for individual generators. \textbf{The higher the better.}  The rankings are overall similar, as GAN-based generators are ranked low.}
  \label{word_overlap_eval_compressed_version}
\end{figure*}

In Figure \ref{word_overlap_eval_compressed_version} we present detailed results for all word overlap evaluators we used in this study.

\subsubsection{Comparing Evaluators}
\label{appendix_comparing_evaluators}

In Table \ref{table_correlation_results} we present correlation results between the evaluators included in this work. 

\begin{table*}[htbp]
\centering
\scalebox{0.8}{
\begin{tabular}{  l | c | c | c | c | c | c }
\hline
\hline
\textbf{Evaluation Method} & \textbf{Kendall tau-b} & \textbf{Spearman} & \textbf{Pearson} & \textbf{Kendall tau-b} & \textbf{Spearman}  & \textbf{Pearson} \\
 & \textbf{(H1)} &  \textbf{(H1)} &  \textbf{(H1)} & \textbf{(H2)} &  \textbf{(H2)} & \textbf{(H2)}  \\
 \hline
\textbf{SVM} Individual-discriminators & -0.4545* & -0.6294* & -0.6716*  & -0.5455* & -0.6783* & -0.6823* \\
\hline
\textbf{LSTM} meta-discriminator & -0.5455* & -0.7552* &  -0.7699* & -0.6364* & -0.8042* & -0.7829* \\
\hline
\textbf{CNN} meta-discriminator & -0.6363* & -0.8112* & -0.8616* & -0.7273* & -0.8741* & -0.8766* \\
\hline
\textbf{CNN \& LSTM} meta-discriminator & -0.6060* &  -0.7902* & -0.8392* & -0.6970* & -0.8462* & -0.8507* \\
\hline
\textbf{SVM} meta-discriminator & -0.4545* & -0.6573* & -0.7207* & -0.5455* & -0.6993* & -0.7405 \\
\hline
\textbf{RF} meta-discriminator & -0.5455* & -0.7273* & -0.7994* & -0.6364* & -0.7832* & -0.8075* \\
\hline
\textbf{NB} meta-discriminator & -0.6364* & -0.8112* & -0.9290* & -0.7273* & -0.8741* & -0.9388* \\
\hline
\textbf{XGBoost} meta-discriminator & -0.5455* & -0.7413* & -0.7764* & -0.6364* & -0.8042* & -0.7878* \\
\hline
\textbf{BLEU} evaluator  & 0.7576* & 0.8601* & 0.8974* & 0.6666* & 0.8182* & 0.9060* \\
\hline
\textbf{ROUGE} evaluator & 0.6060* & 0.7692* & 0.8054* & 0.5758* & 0.7483* & 0.8073* \\
\hline
\textbf{METEOR} evaluator & 0.5758* & 0.7762* & 0.8225* & 0.5455* & 0.7622* & 0.8231* \\
\hline
\textbf{CIDEr} evaluator & 0.5455* & 0.7413* & 0.8117* & 0.4545* & 0.6643* & 0.8203* \\ 
\hline
\hline
\end{tabular}}
\caption{Kendall tau-b, Spearman and Pearson correlation coefficients between human evaluators $H1$, $H2$, and discriminative evaluators and word-overlap evaluators (* denotes statistical significant result with $p \le 0.05$).} 
\label{table_correlation_results}
\end{table*}

\subsubsection{Diversity Analysis}
\label{appendix_diversity_analysis}

In Table \ref{diversity_results} we present results for the Self-BLEU metric, while in Table \ref{table_correlation_diversity} we present the correlation of Self-BLEU with the other evaluators. In addition, in Table \ref{table_correlation_BLEUGTrain} we present correlation results for BLEU G-Train and the rest of the evaluators.

\begin{table}[!htbp]
\centering
\scalebox{0.7}{
\begin{tabular}{  l | c | c }
\hline
\hline
\textbf{Generative Text Model} & \textbf{Self-BLEU} & \textbf{Lexical diversity}\\
\hline
Word LSTM temp 1.0 & 0.1886 & 0.6467 \\
\hline
Word LSTM temp 0.7 & 0.4804 & 0.2932 \\
\hline
Word LSTM temp 0.5 & 0.6960 & 0.1347 \\
\hline
Scheduled Sampling & 0.1233 & 0.7652 \\
 \hline
Google LM & 0.1706 & \textbf{0.7745} \\
\hline
Attention Attribute to Sequence & 0.5021 & 0.2939 \\
 \hline
Contexts to Sequences & 0.8950 & 0.0032 \\
 \hline
Gated Contexts to Sequences & 0.7330 & 0.1129 \\
 \hline
MLE SeqGAN & 0.1206 & 0.7622 \\
 \hline
SeqGAN & 0.1370 & 0.7330 \\
 \hline
RankGAN & \textbf{0.1195} & 0.7519 \\
 \hline
LeakGAN & 0.1775 & 0.7541 \\
\hline
\hline
\end{tabular}}
\caption{Self-BLEU diversity scores per generator (the lower the more diverse), and lexical diversity scores (the higher the more diverse). There is high correlation between the two metrics with respect to the rankings of the generative text models.}
\label{diversity_results}
\end{table}

\begin{table}[!htbp]
\centering
\scalebox{0.6}{
\begin{tabular}{  l | c | c | c }
\hline
\hline
\textbf{Self-BLEU} & \textbf{Kendall tau-b} & \textbf{Spearman} & \textbf{Pearson} \\
\hline
\textbf{H1} evaluator & -0.8788* & -0.9301* & -0.8920* \\
\hline
\textbf{H2} evaluator & -0.7879* & -0.8881* & -0.9001* \\
\hline
\textbf{LSTM} meta-discriminator & 0.6667* & 0.8252* & 0.7953* \\
\hline
\textbf{CNN} meta-discriminator & 0.7576* & 0.8811* & 0.8740* \\
\hline
\textbf{CNN \& LSTM} meta-discriminator & 0.7273* & 0.8601*
& 0.8622* \\
\hline
\textbf{SVM} meta-discriminator & 0.5758* & 0.7413* & 0.8518* \\
\hline
\textbf{RF} meta-discriminator & 0.6667* & 0.8112* & 0.8944* \\
\hline
\textbf{NB} meta-discriminator & 0.7576* & 0.8811* & 0.9569* \\
\hline
\textbf{XGBoost} meta-discriminator & 0.6667* & 0.8252*
 & 0.8693* \\
\hline
\textbf{BLEU} evaluator & -0.8788 & -0.9301* & -0.9880* \\
\hline
\textbf{ROUGE} evaluator & -0.7273* & -0.8392* & -0.9299* \\
\hline
\textbf{METEOR} evaluator & -0.6967* & -0.8462* & -0.8955* \\
\hline
\textbf{CIDEr} evaluator & -0.5455* & -0.7413* & -0.7987*\\
\hline
\hline
\end{tabular}}
\caption{Kendall tau-b, Spearman and Pearson correlation coefficients between Self-BLEU diversity rankings and the three evaluation methods - human evaluators $H1$, $H2$, discriminative evaluators and word-overlap based evaluators (* denotes statistical significant result with $p \le 0.05$). Meta-discriminators have been trained on D-train, D-valid sets and tested on the \textbf{annotated D-test set with ground-truth test labels}.} 
\label{table_correlation_diversity}
\end{table}

\begin{table}[t!]
\centering
\scalebox{0.6}{
\begin{tabular}{  l | c | c | c }
\hline
\hline
\textbf{BLEU G-train} & \textbf{Kendall tau-b} & \textbf{Spearman} & \textbf{Pearson} \\
\hline
\textbf{H1} evaluator & 0.7176* & 0.8511* & 0.9111* \\
\hline
\textbf{H2} evaluator & 0.6260* & 0.8091* & 0.9209* \\
\hline
\textbf{LSTM} meta-discriminator & -0.5649* & -0.7461* & -0.7091* \\
\hline
\textbf{CNN} meta-discriminator & -0.6565 & -0.7951* & -0.8213* \\
\hline
\textbf{CNN \& LSTM} meta-discriminator & -0.6260* & -0.7811* & -0.7951* \\
\hline
\textbf{SVM} meta-discriminator & -0.4428* & -0.6130* & -0.7442*\\
\hline
\textbf{RF} meta-discriminator & -0.5038* & -0.6340* & -0.7864*\\
\hline
\textbf{NB} meta-discriminator & -0.6260* & -0.7601* & -0.9164* \\
\hline
\textbf{XGBoost} meta-discriminator & -0.5649* & -0.6550* & -0.7586*\\
\hline
\textbf{BLEU} evaluator & 0.9619* & 0.9912* & 0.9936* \\
\hline
\textbf{ROUGE} evaluator & 0.5954* & 
0.7496* & 0.8717* \\
\hline
\textbf{METEOR} evaluator & 0.6260* & 
0.7636* & 0.8477* \\
\hline
\textbf{CIDEr} evaluator & 0.6565* & 0.8371* & 0.8318* \\
\hline
\hline
\end{tabular}}
\caption{Kendall tau-b, Spearman and Pearson correlation coefficients between BLEU G-train rankings and the three evaluation methods - human evaluators $H1$, $H2$, discriminative evaluators and word-overlap based evaluators (* denotes statistical significant result with $p \le 0.05$). Meta-discriminators have been trained on D-train, D-valid sets and tested on the \textbf{annotated D-test set with ground-truth test labels}.} 
\label{table_correlation_BLEUGTrain}
\end{table}

\section{Discussion}

\subsection{User Study}
\label{appendix_user_study}

A more detailed list of major clusters of reasons is as follows: 

\begin{enumerate}
\item Grammar/ typo/ mis-spelling: the language does not flow well.
\item Too general/ too generic/ vagueness: generated reviews are vague, in lack of details.
\item Word choice (wording): in lack of slang, use the wrong words.
\item Flow (not fluent)/ structured/ logical: the sentences level language errors.
\item Contradictory arguments: some arguments support opposite opinions.
\item Emotion: lack of emotion, personality in the comments.
\item Repeated text: using words/ phrases repetitively.
\item Overly same as human: too  advertisement, too formal, too likely to be real.
\end{enumerate}

\subsection{Granularity of Judgements}
\label{appendix_granularity_of_judgments}

We charged the Turkers to label individual reviews as either fake or real. Each human judge only annotates 20 reviews, and they do not know which reviews are generated by the same generator.  
Comparing to an adversarial discriminator, a human judge has not seen many ``training'' examples of fake reviews or generators.  That explains why the meta-adversarial evaluators are better at identifying fake reviews.  In this context, humans are likely to judge whether a review is real based on how ``similar'' it appears to the true reviews they are used to see online.  That is probably why their decisions are better correlated to text-overlap metrics that measures the similarity between a review and a set of references.  This hypothesis is supported by a post-experiment survey of the human judges. Please see Appendix \ref{appendix_user_study_samples} for user study samples. 

This finding provides interesting implications to the selection of evaluation methods for different tasks.  In tasks that are set up to judge individual pieces of generated text (e.g., reviews, translations, summaries, captions, fake news) where there exists human-written ground-truth, it is better to use word-overlap metrics instead of adversarial evaluators.  Indeed, when the audience are not trained by reading lots of bot-generated texts,  it is more reasonable to use an evaluator that mimics their decision-making process. 

In some scenarios, the task is to make judgments in the context of a longer conversation or a set of documents (e.g., conversation agents, dialogue systems, social bots).  The difference is that human subjects are exposed to machine-generated text, so that they may be better trained to distinguish fake from real.  Moreover, when judgments are made on the agent/ system level (e.g., whether a Twitter account is a bot), signals like how similar the agent outputs are or how much the agent memorizes the training examples may become more useful than word usage, and a discriminative evaluator may be more effective than text-overlap metrics. 

Our experiment also provide implications to improving NLG models, which implies that adversarial accuracy might not be the optimal objective for NLG if the goal is to generate documents that humans consider as real.  Indeed, a fake review that fools humans does not necessarily need to fool a machine that has seen everything.  

In contrast, GAN based models may perform better when judged as a whole system instead of individual items, or in a conversational context.  When the human judges have seen enough examples from the same generator, the next example had better be somewhat different.

\subsection{Imperfect Ground-truth}
\label{appendix_imperfect_ground_truth}

One important thing to note is that all discriminative evaluators are trained using natural labels (i.e., treating all examples from the Amazon review dataset as positive and examples generated by the candidate models as negative) instead of human-annotated labels.  It is possible that if they were trained with human labels, the discriminative evaluators would have been more consistent to the human evaluators.  Indeed, some reviews posted on Amazon may have been generated by bots, and if that is the case, treating them as human-written examples may bias the discriminators.  



One way to verify this is to consider an alternative ``ground-truth''. 
We apply the already trained meta-discriminators 
to the human-annotated subset (3,600 reviews) instead of the full \textit{D-test} set, and we use the majority vote of human judges (whether a review is fake or real) to surrogate the ``ground-truth'' labels (whether a review is generated or sampled from Amazon). 

\begin{figure}[!htbp]
\centering
  \includegraphics[width=2.5in]{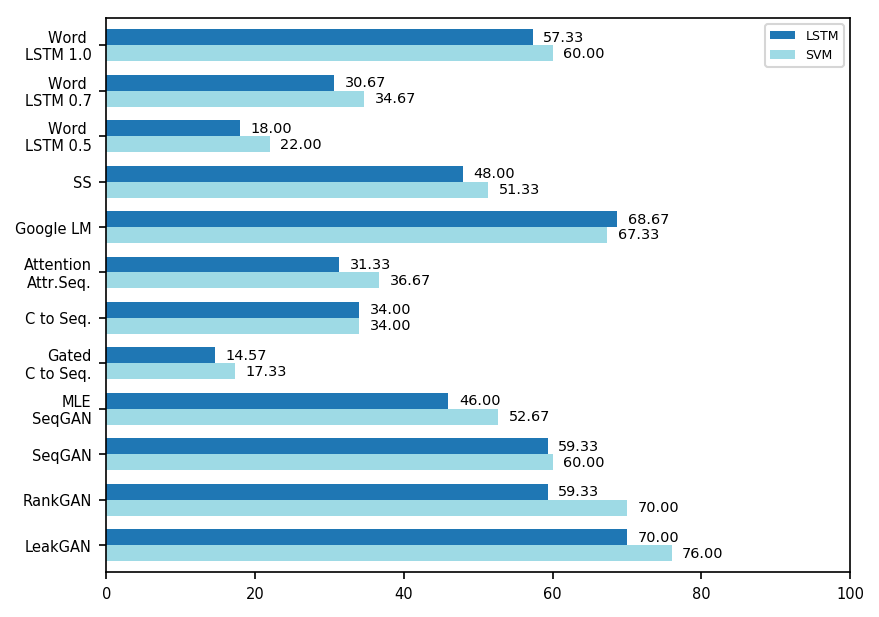}
  \caption{Accuracy of deep (LSTM) and shallow (SVM) meta-discriminators when tested on the \textbf{annotated subset of \textit{D-test}}, with \textit{majority votes} as ground-truth. The lower the better.}
  \label{fig::heatmap_annotated_Dtest_majority_vote_test_labels}
\end{figure}

Surprisingly, when the meta-adversarial evaluators are tested using human majority-votes as ground-truth, both the accuracy numbers and the rankings of the generators are significantly different from Table~\ref{meta_discriminator_accuracy_rankings} and Table~\ref{meta_discriminator_accuracy_rankings_all} (which used natural labels as ground-truth). We note that the scores and rankings are more inline with the human evaluators. 
To confirm the intuition, we calculate the correlations between the meta-discriminators and the human evaluators using the annotated subset only. 
Replacing the natural ground-truth with human annotated labels, the meta-discriminators become positively correlated with human evaluators (Figure~\ref{fig::hbarchart_annotatedDtest_majority_vote_test_labels}), although BLEU still appears to be the best evaluator. 

These results indicate that when the ``ground-truth'' used by an automated Turing test is questionable, the decisions of the evaluators may be biased.  Discriminative evaluators suffer the most from the bias, as they were directly trained using the imperfect ground-truth.  Text-overlap evaluators are more robust, as they only take the most relevant parts of the test set as references (more likely to be high quality).  

Our results also suggest that when adversarial training is used, the selection of training examples must be done with caution.  If the ``ground-truth'' is hijacked by low quality or ``fake'' examples, models trained by GAN may be significantly biased.  This finding is related to the recent literature of the robustness and security of machine learning models.

\subsection{Role of Diversity}
\label{appendix_role_of_diversity}

We also assess the role diversity plays in the rankings of the generators. To this end, we measure lexical diversity \cite{bache2013text} of the samples produced by each generator as the ratio of unique tokens to the total number of tokens.  We compute in turn lexical diversity for unigrams, bigrams and trigrams, and observe that the generators that produce the least diverse samples are easily distinguished by the meta-discriminators, while they confuse human evaluators the most.  Alternatively, samples produced by the most diverse generators are hardest to distinguish by the meta-discriminators, while human evaluators present higher accuracy at classifying them.  As reported in \cite{kannan2017adversarial}, the lack of lexical richness can be a weakness of the generators, making them easily detected by a machine learning classifier.  Meanwhile, a discriminator's preference for rarer language does not necessarily mean it is favouring higher quality reviews.

\begin{figure}[!htbp]
\centering
  \includegraphics[width=\columnwidth]{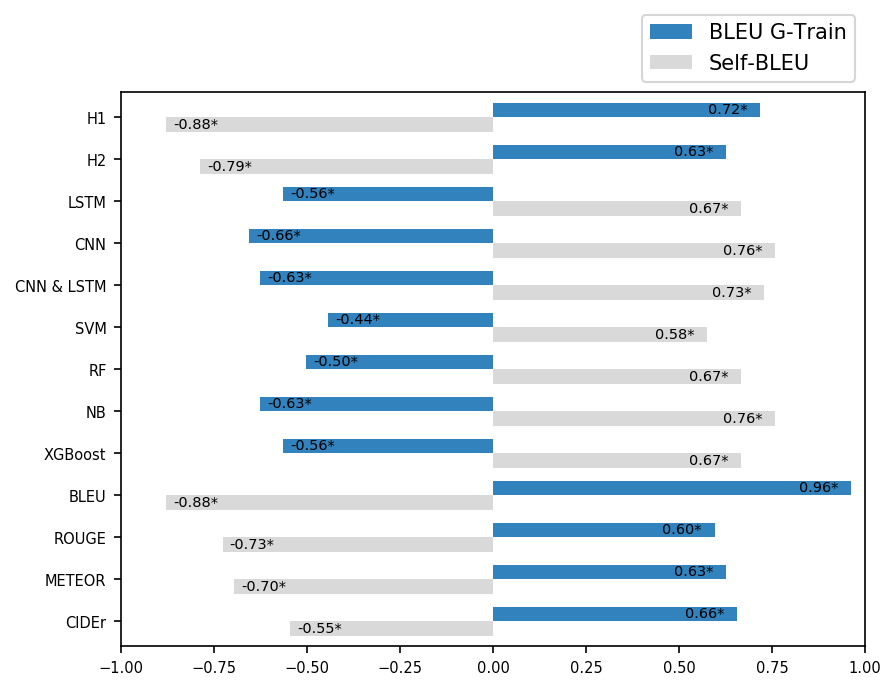}
  \caption{Kendall $\tau$-b correlation coefficients between BLEU G-train and Self-BLEU rankings, and the three evaluation methods - human evaluators $H1$, $H2$, discriminative evaluators and word-overlap based evaluators (* denotes $p \le 0.05$). Meta-discriminators have been trained on D-train, D-valid sets and tested on the \textbf{annotated D-test set with ground-truth test labels}.}
  \label{fig::correlation_SelfBLEU_BLEUGTrain}
\end{figure}

In addition to lexical diversity, Self-BLEU \cite{zhu2018texygen} is an interesting measurement of the diversity of a set of text (average BLEU score of each document using the same collection as reference, therefore the lower the more diverse). In 
Figure \ref{fig::self_bleu_lexical_diversity} we present Self-BLEU scores for each generator, applied to their generated text in \textit{D-test fake}.  We also compute the correlation coefficients between the rankings of generators by Self-BLEU and the rankings by the evaluators (please see Figure \ref{fig::correlation_SelfBLEU_BLEUGTrain}). 
Results obtained indicate that Self-BLEU presents negative correlation with human evaluators and word-overlap evaluators, and positive correlation with discriminative evaluators.  This result confirms the findings in literature  \cite{kannan2017adversarial} that discriminators in adversarial evaluation are capturing known limitations of the generative models such as lack of diversity. 

Following this insight, an important question to answer is to what extent the generators are simply memorizing the training set \textit{G-train}. To this end, we assess the degree of n-gram overlap between the generated reviews and the training reviews using the BLEU evaluator. In Table \ref{results_BLEUGTrain} we present the average BLEU scores of generated reviews using their nearest neighbors in \textit{G-train} as references.  We observe that generally the generators do not memorize the training set, and GAN models generate reviews that have fewer overlap with \textit{G-train}. In  
Figure \ref{fig::correlation_SelfBLEU_BLEUGTrain} 
we include the correlation between the divergence from training and the ratings by evaluators in the study. BLEU w.r.t. \textit{G-train} presents highly positive correlation with BLEU w.r.t. \textit{D-test real}, and it is also positively correlated with the human evaluators $H1$ and $H2$. 

\begin{table}[t!]
\centering
\scalebox{0.7}{
\begin{tabular}{  l | c  }
\hline
\hline
\textbf{Generative Text Model} & \textbf{BLEU G-Train} \\
\hline
Word LSTM temp 1.0 & 0.2701 \\
\hline
Word LSTM temp 0.7 & 0.4998 \\
\hline
Word LSTM temp 0.5 & 0.6294 \\
\hline
Scheduled Sampling & 0.1707 \\
 \hline
Google LM & 0.0475 \\
\hline
Attention Attribute to Sequence & 0.5122 \\
 \hline
Contexts to Sequences & 0.7542 \\
 \hline
Gated Contexts to Sequences & 0.6240 \\
 \hline
MLE SeqGAN & 0.1707 \\
 \hline
SeqGAN & 0.1751 \\
 \hline
RankGAN & 0.1525 \\
 \hline
LeakGAN & 0.1871 \\
\hline
\hline
\end{tabular}}
\caption{BLEU results when evaluating the generated reviews using G-train as the reference corpus (a lower score indicates less n-grams in common between the training set G-train and the generated text). GAN models present low similarity with the training set.}
\label{results_BLEUGTrain}
\end{table}

The effects of diversity is perhaps not hard to explain. At the particular task of distinguishing fake reviews from real, all decisions are made on individual reviews. And because a human judge was not exposed to many fake reviews generated by the same generator, whether or not a fake review is sufficiently different from the other generated reviews is not a major factor for their decision.  Instead, the major factor is whether the generated review looks similar to the reviews they have seen in reality.  Instead, a discriminative evaluator makes the decision after seeing many positive and negative examples, and a fake review that can fool an adversarial classifier has to be sufficiently different from all other fake reviews it has encountered (therefore diversity of a generator is a major indicator of its ability to pass an adversarial judge).